\pdfoutput=1

\documentclass[10pt,twocolumn,letterpaper]{article}

\usepackage[pagenumbers]{cvpr}

\usepackage{placeins}

\definecolor{cvprblue}{rgb}{0.21,0.49,0.74}
\usepackage[breaklinks,colorlinks,allcolors=cvprblue,hypertexnames=false]{hyperref}
\usepackage{orcidlink}

\def\paperID{*****} 
\def\confName{CVPR}
\def\confYear{2027}

\title{ApexID: Hierarchical Learning for Multi-Species Animal Re-Identification}

\author{
Seifeldin Elkerdany$^{1,2,*,\dagger}$\,\orcidlink{0009-0005-7199-7391}
\quad
Mohamed ElBassat$^{1,3,*,\dagger}$\,\orcidlink{0009-0006-3917-1319}
\quad
Gamal Abouelhamd$^{1,2}$\,\orcidlink{0009-0004-7223-5260}
\\
Assem Elkady$^{1,3}$\,\orcidlink{0009-0003-9847-9944}
\quad
Jana Ghoneim$^{1,3}$\,\orcidlink{0009-0008-9943-295X}
\quad
Matin Mahmood$^{4}$
\quad
Abigail Allen Martin$^{5}$\,\orcidlink{0009-0001-6233-2735}
\\
Gerard de Melo$^{6}$\,\orcidlink{0000-0002-2930-2059}
\quad
Antonio Rueda-Toicen$^{6}$\,\orcidlink{0000-0002-6812-7337}
\\[0.8em]
{\small
$^{1}$Made In Alexandria AI Team (M.I.A Robotics)
\quad
$^{2}$Alamein International University
\quad
$^{3}$Alexandria University}
\\
{\small
$^{4}$hyper\textsuperscript{3}labs
\quad
$^{5}$Universidade Federal de Mato Grosso
\quad
$^{6}$Hasso Plattner Institute}
\\[0.4em]
{\small $^{*}$Equal contribution
\quad
$^{\dagger}$Corresponding authors}
\\
{\tt\small seifeldean.hosny.2024@aiu.edu.eg, cds.MohamedMohamed24510@alexu.edu.eg}
}

\begin{document}
\maketitle

\begin{abstract}
Animal re-identification (Re-ID) must distinguish individuals across
species and capture conditions. We introduce ApexID, a hierarchical
SphereFace2-based loss that
integrates global identity discrimination with balanced animal-category
supervision.
Using 132,533 images curated from 19 public datasets, we evaluate three
model families and four objectives on four external datasets with
observation-aware retrieval protocols. ApexID achieves a dataset-averaged
Rank-1 accuracy of 64.89\% with ConvNeXt-Small and an mAP of 32.54\%
with MIEW-ID, outperforming their pretrained reference models.
On elephants, a category absent from animal Re-ID training, our
ConvNeXt-Small variant achieves the highest retrieval performance among
the fourteen evaluated configurations. Scores vary by backbone and
dataset, and open-set evaluation reveals remaining discovery challenges.
These findings support hierarchical category--identity learning to
improve transfer to unseen datasets and species.
Code, weights, and data splits are available in our
\href{https://github.com/MIA-AI-Team/ApexID-Hierarchical-Learning-for-Multi-Species-Animal-Re-Identification-Official-Repo}{GitHub repository}.
\end{abstract}    
\section{Introduction}
\label{sec:intro}

Individual animal re-identification (Re-ID) supports non-invasive population
monitoring, movement analysis, and behavioral research by matching individuals
across sightings~\cite{tuia2022perspectives,vcermak2024wildlifedatasets}.
Reliable matching must preserve subtle individual cues despite changes in
pose, viewpoint, illumination, and time. A representation shared across
animal categories must also accommodate broad appearance differences while
distinguishing individuals within each category.

Standard identity-only objectives do not explicitly model the category--identity
relationships present in multi-species annotations. Meanwhile, backgrounds
and acquisition conditions may correlate with identity or dataset, creating
contextual cues that may not persist across sightings. These challenges
motivate category-aware identity learning from curated animal foregrounds.

We introduce \textbf{ApexID}, a multi-species animal Re-ID approach that maps
curated animal foregrounds into a shared embedding space. We train its
encoders with the \emph{ApexID Hierarchical Loss}, a two-level
category--identity objective based on SphereFace2~\cite{wen2022sphereface2}.
The loss combines global identity discrimination over learnable identity
sub-centers with balanced category supervision. For hierarchical supervision, an identity's parent is a broad
animal category, such as whale. Once the foregrounds have been extracted,
closed-set retrieval ranks gallery images by cosine similarity between
normalized embeddings alone.

We harmonize animal category labels across 19 public datasets, apply
dataset-specific near-duplicate filtering within identities, and segment
the animal foregrounds. We manually review the images, create or correct
masks where needed, and retain 132,533 images for training and validation.

We compare four objectives on three backbones and include two unmodified
animal Re-ID checkpoints as references. Across four external datasets,
ConvNeXt-Small trained with our loss achieves the highest dataset-macro
Rank-1 accuracy (64.89\%) among the 14 configurations and a dataset-macro
mean average precision (mAP) of 30.84\%. Both scores exceed those of the
unmodified MIEW-ID v3 and MegaDescriptor-L checkpoints. MIEW-ID v3
fine-tuned with our loss achieves the highest dataset-macro mAP (32.54\%),
compared with 31.89\% for the same backbone fine-tuned with SphereFace2.
Gains vary by backbone and dataset. We also evaluate the embeddings in
a separate, dataset-adapted open-set protocol that uses calibrated
similarities to associate queries with known identities and group
observations of novel individuals.

Our contributions are threefold:
\begin{itemize}
    \item \textbf{ApexID models for multi-species Re-ID.}
    We fine-tune ImageNet-pretrained ConvNeXt-Small and two animal Re-ID
    encoders, MIEW-ID v3 and MegaDescriptor-L, on curated foregrounds
    using the proposed loss.

    \item \textbf{A hierarchical category--identity objective.}
    \emph{ApexID Hierarchical Loss} augments global identity
    discrimination with balanced category-level supervision through
    a separate learned proxy for each animal category.

    \item \textbf{A curated multi-source dataset.}
    We harmonize annotations from 19 public datasets, remove unusable
    images, and retain 13,348 source-specific identities. We retain
    the source dataset and observation-group identifier for each image.
    Motivated by prior work on time- and similarity-aware
    splitting~\cite{Adam_2025_CVPR}, we assign each observation group
    wholly to either training or validation.
\end{itemize}
\section{Related Work}
\label{sec:related_work}

\paragraph{Shared animal Re-ID representations.}
WildlifeDatasets introduced MegaDescriptor and a unified framework for
learning across heterogeneous wildlife datasets~\cite{vcermak2024wildlifedatasets}.
MIEW-ID studies multi-species animal Re-ID~\cite{otarashvili2024miewid}, while
OpenAnimals investigates the transfer of person Re-ID techniques to
animals~\cite{hou2025openanimals}. We evaluate category-aware supervision
across MIEW-ID v3, MegaDescriptor-L, and ImageNet-pretrained ConvNeXt-Small.

\paragraph{Recognition objectives.}
ArcFace applies a fixed additive angular margin to the target class~\cite{deng2019arcface}, while Dynamic ArcFace adapts this margin during training~\cite{jiao2021dynarcface}. Sub-center ArcFace represents each identity with multiple learned centers~\cite{deng2020subcenter}, and SphereFace2 trains one binary classifier per identity~\cite{wen2022sphereface2}. Yang et al.~\cite{yang2022hierarchical} construct coarse proxies by clustering class proxies and also evaluate a hierarchy defined by annotations.
The \emph{ApexID Hierarchical Loss} uses the annotated animal category as an additional prediction target, with one learned proxy per category, alongside sub-center SphereFace2 identity discrimination.

\paragraph{Foregrounds and observation-aware evaluation.}
Rueda-Toicen et al.~\cite{ruedatoicen2026jaguar} examine background reliance and
laterality in jaguar Re-ID, illustrating why retrieval scores alone do not
identify the visual evidence driving a match. ApexID applies foreground
processing across multiple animal categories. WildlifeReID-10k uses timestamps when available and visual similarity otherwise to reduce overlap between partitions~\cite{Adam_2025_CVPR}. We follow this principle by keeping observation groups intact in both training--validation and closed-set gallery--query splits. For our separate batch open-set analysis, we adapt the calibrated similarity and graph clustering used by ElBassat et al.~\cite{elbassat2026calibrated} in their AnimalCLEF~2026 system.

\section{Dataset Construction and Handling}
\label{sec:dataset}

\subsection{Sources and Scope}
\label{sec:dataset:sources}
ApexID combines 19 public animal Re-ID datasets that vary in annotation,
image quality, and observation metadata. For each image, we retain its source dataset, animal category, individual identity, and observation group. Each individual is identified by the combination of its source dataset and its local identity label. Animal categories represent broad categories shared across the datasets. 

Table~\ref{tab:apexid_corpus_summary}
summarizes the curated corpus, and Supplementary
Table~\ref{tab:apexid_train_datasets} lists all sources.

\begin{table}[!htbp]
\centering
\caption{Primary ApexID training and validation partitions. Identities are counted within each source dataset. Known-validation identities also appear in training through different observation groups.}
\label{tab:apexid_corpus_summary}
\small
\setlength{\tabcolsep}{5pt}
\begin{tabular}{@{}l r r@{}}
\toprule
\textbf{Partition} & \textbf{Images} & \textbf{Identities} \\
\midrule
Training & 90,921 & 11,677 \\
Known-identity validation & 28,070 & 10,768 \\
Held-out-identity validation & 13,542 & 1,671 \\
\midrule
Total (unique) & 132,533 & 13,348 \\
\bottomrule
\end{tabular}
\end{table}

The primary partitions contain 96,435 observation groups. Within each source-specific identity, these groups are constructed from capture dates, verified filename or session information, or visual similarity, depending on the metadata available. Together, the two validation
partitions contain 41,612 images across 12,439 identities. We exclude all four external evaluation datasets from ApexID training and checkpoint selection. Sec.~\ref{sec:experiments:datasets} describes their roles in closed-set retrieval and dataset-adapted open-set evaluation.

\subsection{Preprocessing and Curation}
\label{sec:dataset:curation}
Before segmentation and partitioning, we use MegaDescriptor embeddings~\cite{vcermak2024wildlifedatasets} to filter near-duplicates in the datasets. We compare images only within the same source-specific identity and use a dataset-specific cosine-similarity threshold.
\label{sec:dataset:duplicates}

We segment images with Segment Anything Model~3 (SAM~3)~\cite{carion2025sam3}
using an animal-category text prompt. We retain the highest-confidence valid
animal mask and set the pixels outside it to black.
\label{sec:dataset:segmentation}
Using V-Rush~\cite{elbassat_vrush_2026}, we manually review images for poor
quality, segmentation failures, and annotation inconsistencies. We generate or
refine incomplete masks with bounding-box-prompted
MobileSAM~\cite{zhang2023mobilesam} and correct them with a manual brush. We
exclude samples whose identity annotations remain unresolved. Counts are computed after
curation; additional procedural details appear in Supplementary
Sec.~\ref{sec:suppl:data}.

\subsection{Observation-Aware Partitioning}
\label{sec:dataset:partitioning}
Following observation-aware evaluation~\cite{Adam_2025_CVPR}, we keep each
observation group in a single partition. Within each source dataset, an
identity is eligible for the split if at least two images are labeled as
the same identified animal. Using seed~42, we select approximately 12.5\%
of these identities for held-out-identity validation and exclude all
their images from training.

For each eligible identity outside the holdout, we target 22.5\% of its
images for known-identity validation while retaining at least one training
group. If all images are dated and span at least two days, we choose the
cutoff that keeps groups intact and comes closest to the target, assigning
earlier groups to training and later groups to validation. Otherwise, if
verified filename or session information groups all images into at least
two groups, we deterministically select whole groups for validation.

If neither rule applies, we group images separately for each animal.
Two images go in the same group if they are exact copies, share a
recorded capture date, come from the same verified capture sequence,
or have a high DINOv2 similarity score~\cite{oquab2023dinov2}. All groups with two or more images
go to training. We select validation images from the single-image
groups and use the rest for training. See Supplementary
Sec.~\ref{sec:suppl:partitioning} for details.


\section{Method}
\label{sec:method}

We present ApexID, a multi-species animal re-identification approach that learns a shared identity-discriminative embedding from heterogeneous wildlife datasets. Its hierarchical objective combines global identity discrimination with balanced animal-category supervision, encouraging identity-level feature learning while incorporating category-level information.

\subsection{Problem Formulation}
\label{sec:method:formulation}

\paragraph{Training labels.}
Our training set is $\mathcal{D}_{\mathrm{tr}}=\{(x_i,y_i,s_i)\}_{i=1}^N$. Each of the $N$ images $x_i$ depicts one of $C$ training individuals. Its identity $y_i\in\{1,\ldots,C\}$ is defined by its source dataset and local identity label. Each image is also assigned an animal category label $s_i\in\{1,\ldots,S\}$ when available. The observed categories include both species-specific labels and broader animal categories, such as \emph{whale}. We write $s_i=\bot$ when the image lacks a usable category label.

For each training identity $j$, we obtain its parent category $h(j)$
from the images of that individual. We reject conflicting category
annotations. For each observed category $s$, we define its set of training
identities $\mathcal{I}_s$ and their number $n_s$ as
\begin{equation}
\mathcal{I}_s=\{j:h(j)=s\}, \qquad n_s=|\mathcal{I}_s|.
\label{eq:formulation:hierarchy}
\end{equation}

\paragraph{Retrieval.}
The encoder $f_\theta$ maps a preprocessed image $x\in\mathcal{X}$ to a
$D$-dimensional vector. Here $\theta$ denotes the learned encoder
parameters. We scale each vector to unit length to obtain the embedding
\begin{equation}
\mathbf{z}(x)=\frac{f_\theta(x)}{\|f_\theta(x)\|_2},
\qquad \mathbf{z}_i=\mathbf{z}(x_i).
\label{eq:formulation:embedding}
\end{equation}
Given a query image $x_q$, we rank the $M$ labeled reference images in the
gallery $\mathcal{G}=\{(x_g,y_g)\}_{g=1}^M$. We rank the gallery images from highest to lowest similarity. For unit-length embeddings, this is their dot product,
\begin{equation}
\operatorname{sim}(x_q,x_g)=\mathbf{z}(x_q)^\top\mathbf{z}(x_g).
\label{eq:formulation:similarity}
\end{equation}

\subsection{Model Architecture}
\label{sec:method:architecture}

We train each backbone separately and use the same encoder for all animal
categories within a model. For ImageNet-pretrained
ConvNeXt-Small~\cite{liu2022convnet}, we replace the final classification
layer with a learned linear projection to $D=2048$. For the two encoders
pretrained on animal Re-ID, we use their native embeddings directly.
The embedding dimension is $2152$ for MIEW-ID
v3~\cite{otarashvili2024miewid} and $1536$ for
MegaDescriptor-L~\cite{vcermak2024wildlifedatasets}.

When training with our loss, we attach two heads to the normalized
embedding. The identity head represents each training individual with $K$
learned reference vectors (proxies), called \emph{sub-centers}. The category head
uses one learned reference vector per category. At
inference, we keep the encoder and discard both training heads.

\subsection{ApexID Hierarchical Loss}
\label{sec:method:loss}

\begin{figure*}[!t]
    \centering
    \includegraphics[width=\textwidth]{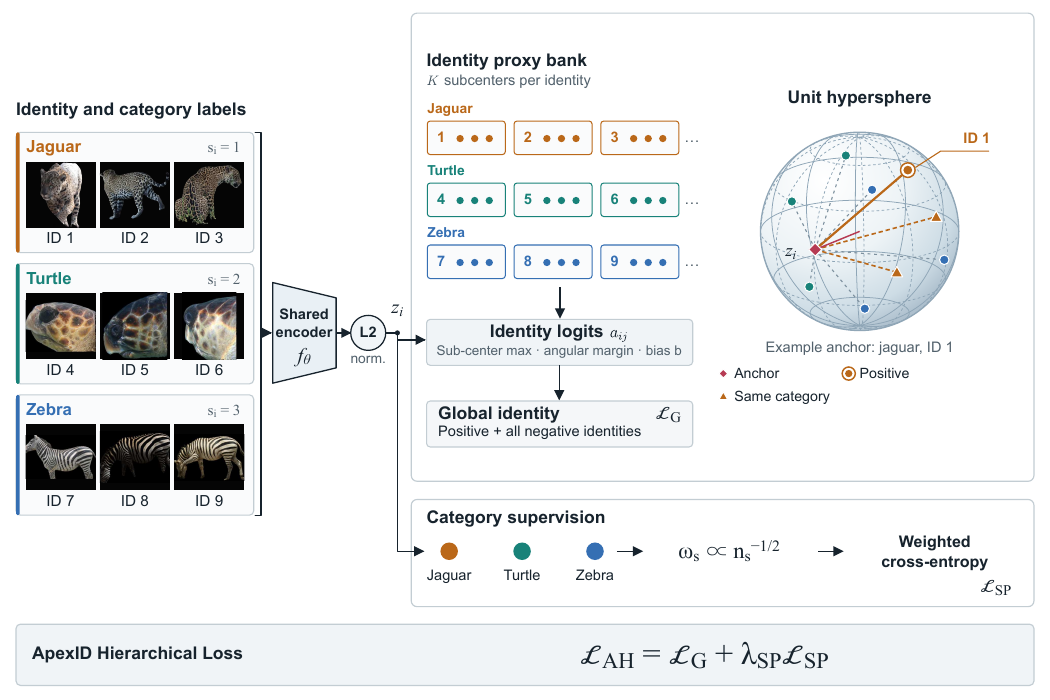}
    \caption{The encoder produces one normalized embedding per image. The global
identity loss compares it with every training identity using learned
identity sub-centers. A separate category prediction head uses
learned category proxies and balanced supervision, assigning greater
weight to categories with fewer training identities.}
    \label{fig:apexid_hierarchical_loss}
\end{figure*}

We build on SphereFace2~\cite{wen2022sphereface2} and hierarchical proxy-based metric learning~\cite{yang2022hierarchical}. As shown in Fig.~\ref{fig:apexid_hierarchical_loss}, the loss combines global identity discrimination ($\mathcal{L}_{\mathrm{G}}$) and category prediction ($\mathcal{L}_{\mathrm{SP}}$) in a single objective,
\begin{equation}
\boxed{\mathcal{L}_{\mathrm{AH}}=\mathcal{L}_{\mathrm{G}}
 +\lambda_{\mathrm{SP}}\mathcal{L}_{\mathrm{SP}}.}
\label{eq:scsf2:total}
\end{equation}
The nonnegative coefficient $\lambda_{\mathrm{SP}}$ controls the contribution of the category prediction task relative to global identity discrimination.

In our experiments, all $B$ images in a batch have category labels, so every image contributes to both the global identity loss and the category prediction loss. Accordingly, the set of valid samples for both objectives is $\mathcal{V}=\{1,\ldots,B\}$.

\paragraph{Shared identity scores.}
For image $i$, we compare its embedding $\mathbf{z}_i$ with the $K$
sub-centers of each training identity $j$. The largest cosine similarity
gives the score
\begin{equation}
c_{ij}=\max_{1\leq k\leq K}\mathbf{z}_i^\top\widehat{\mathbf{w}}_{j,k}.
\label{eq:scsf2:subcenter}
\end{equation}
Here $\mathbf{w}_{j,k}$ is the $k$th sub-center of identity $j$, and a
hat denotes a vector scaled to unit length. Different photos of the same
individual can therefore match different sub-centers.

For image $i$ and training identity $j$, $c_{ij}$ is the cosine
similarity to the closest sub-center of identity $j$
(Eq.~\eqref{eq:scsf2:subcenter}). Following
SphereFace2-A~\cite[App.~F]{wen2022sphereface2}, we convert this
similarity to an angle $\theta_{ij}=\arccos(c_{ij})$. We then add
an angular margin $m_{\mathrm{ID}}$, measured in radians, only when
$j$ is the image's true identity $y_i$. The adjusted angle is
\[
\tilde{\theta}_{ij} =
\begin{cases}
\theta_{ij}+m_{\mathrm{ID}}, & j=y_i,\\
\theta_{ij},                 & j\ne y_i.
\end{cases}
\]
Using the adjusted angle, we compute the training
score, or logit, for identity $j$ as
\begin{equation}
a_{ij}
= r\,g_t(\cos\tilde{\theta}_{ij})+b,
\qquad
g_t(u)=2\left(\frac{1+u}{2}\right)^t-1.
\label{eq:scsf2:identity_logit}
\end{equation}

The exponent $t>0$ controls how the cosine similarity is transformed,
$r>0$ scales the result, and $b$ is a learned bias shared by all
identities. 

\paragraph{Global identity discrimination.}
Each image has one true identity and $C-1$ other training identities.
For the true identity, the penalty $\ell_i^+$ decreases as its score
rises. For any other identity $j$, the penalty $\ell_{ij}^-$ decreases
as its score falls. These penalties are
\begin{equation*}
\ell_i^+=\log(1+e^{-a_{i y_i}}), \qquad
\ell_{ij}^-=\log(1+e^{a_{ij}}).
\end{equation*}
We sum the penalties for each image and average over the batch to obtain
\begin{equation}
\mathcal{L}_{\mathrm{G}}=\frac{\eta}{rB}\sum_{i=1}^{B}
\left[\alpha\ell_i^+ +(1-\alpha)\sum_{j\ne y_i}\ell_{ij}^-\right].
\label{eq:scsf2:global}
\end{equation}
The coefficient $\alpha\in(0,1)$ weights the true-identity penalty,
$1-\alpha$ weights each negative penalty, and $\eta>0$ scales both
identity losses.


\paragraph{Balanced category supervision.}
For an image with a category label, we ask the model to predict that
category. Each category $s$ has its own learned proxy $\mathbf{v}_s$.
Its cosine similarity to the image is
$q_{is}=\mathbf{z}_i^\top\widehat{\mathbf{v}}_s$. We apply an ArcFace-style
angular margin~\cite{deng2019arcface} only to the score for the true
category. With margin $m_{\mathrm{SP}}$ in radians and score scale
$\gamma>0$, the category logits are
\begin{equation}
u_{is}=
\begin{cases}
\gamma\,\phi_{m_{\mathrm{SP}}}(q_{is}), & s=s_i,\\
\gamma\,q_{is}, & s\ne s_i .
\end{cases}
\label{eq:scsf2:specieslogits}
\end{equation}
Here $\phi_m$ applies the angular margin to the cosine. Its exact form
is given in Eq.~\eqref{eq:scsf2:categorymargin}.

Sampling identities uniformly favors categories that contain more
training individuals. To reduce this imbalance in the category loss,
we give greater weight to categories with fewer training identities,
\begin{equation}
\omega_s=\frac{n_s^{-1/2}}{S^{-1}\sum_{v=1}^{S}n_v^{-1/2}} .
\label{eq:scsf2:weights}
\end{equation}
Let $P_i(s)=\exp(u_{is})/\sum_{v=1}^{S}\exp(u_{iv})$ be the predicted
probability of category $s$. We minimize the weighted cross-entropy over
images with category labels,
\begin{equation}
\mathcal{L}_{\mathrm{SP}}=
-\frac{\sum_{i\in\mathcal{V}}\omega_{s_i}\log P_i(s_i)}
       {\sum_{i\in\mathcal{V}}\omega_{s_i}} .
\label{eq:scsf2:species}
\end{equation}

\paragraph{Margin implementation.}
For the identity logit in Eq.~\eqref{eq:scsf2:identity_logit}, we clip
$c_{ij}$ to $[-1+\delta,1-\delta]$ before computing
$\theta_{ij}=\arccos(c_{ij})$, with $\delta=10^{-5}$. We obtain
$\widetilde{\theta}_{ij}$ by clipping
$\theta_{ij}+m_{\mathrm{ID}}\mathbf{1}_{\{j=y_i\}}$ to $[\delta,3.14159]$.
During backpropagation, we treat the adjustment
$g_t(\cos\widetilde{\theta}_{ij})-c_{ij}$ as constant, giving
$\partial a_{ij}/\partial c_{ij}=r$. This follows the official SphereFace2
implementation~\cite{wen2022sphereface2}.

For the category logit in Eq.~\eqref{eq:scsf2:specieslogits}, we clip
the cosine $q$ to $[-1,1]$. With $m=m_{\mathrm{SP}}$ and $\epsilon=10^{-7}$,
we evaluate the margin function as
\begin{equation}
\begin{aligned}
p_m(q)&=q\cos m-\sqrt{\max(1-q^2,\epsilon)}\sin m,\\
\phi_m(q)&=
\begin{cases}
p_m(q), & q>\cos(\pi-m),\\
q-m\sin(\pi-m), & \text{otherwise}.
\end{cases}
\end{aligned}
\label{eq:scsf2:categorymargin}
\end{equation}
Supplementary Sec. ~\ref{sec:suppl:implementation} gives the hyperparameter values.

\section{Experimental Setup}

\label{sec:experiments}
We compare backbones and loss functions on closed-set and open-set
re-identification across four external datasets.

\subsection{Evaluation Datasets and Model Comparisons}
\label{sec:experiments:datasets}
\label{sec:dataset:test}
We evaluate on ELPephants, GCN-ID, BristolGorillas2020, and private jaguar
data (Table~\ref{tab:apexid_test_datasets}). We apply foreground processing
and manual inspection to all four sources. The datasets and their evaluation
identities are excluded from ApexID representation training and checkpoint
selection, as well as from the animal Re-ID pretraining data for the reference
checkpoints.

\begin{table}[!htbp]
\centering
\caption{External evaluation subsets after preprocessing and selection.} 

\label{tab:apexid_test_datasets}
\scriptsize
\setlength{\tabcolsep}{2pt}
\begin{tabular}{@{}p{0.90in} p{0.76in} r r r@{}}
\toprule
\textbf{Dataset} & \textbf{Species} & \textbf{IDs} &
\textbf{Images} & \textbf{Obs.} \\
\midrule
ELPephants~\cite{korschens2019elpephants} & African forest elephant & 272 & 2,049 & 2,028 \\
GCN-ID~\cite{rosa2025gcnid} & Great crested newt & 207 & 1,232 & 1,152 \\
Bristol Gorillas~\cite{brookes2021bristolgorillas2020,brookes2021gorillas} & Western lowland gorilla & 7 & 5,324 & 4,215 \\
Private jaguar data & Jaguar & 51 & 1,029 & 672 \\
\midrule
\textbf{Total} & -- & \textbf{537} & \textbf{9,634} & \textbf{8,067} \\
\bottomrule
\end{tabular}
\end{table}

\paragraph{Model comparisons.}
\label{sec:experiments:models}
We compare ConvNeXt-Small, MIEW-ID v3~\cite{otarashvili2024miewid}, and
MegaDescriptor-L~\cite{vcermak2024wildlifedatasets}, each trained or fine-tuned
with ArcFace~\cite{deng2019arcface}, Dynamic ArcFace~\cite{jiao2021dynarcface}, SphereFace2~\cite{wen2022sphereface2},
or ApexID Hierarchical Loss. ConvNeXt-Small starts from ImageNet weights; MIEW-ID and
MegaDescriptor-L start from animal Re-ID checkpoints. This yields twelve trained variants and two unmodified animal Re-ID
references, for fourteen configurations in total. Comparisons within each
backbone assess the objectives; comparisons with the original checkpoints
reflect both ApexID fine-tuning and its associated training choices.

\paragraph{Generalization setting.}
All four test datasets evaluate transfer to unseen Re-ID datasets and identities. 
ELPephants also tests an animal category absent from both ApexID
training and the training of MIEW-ID v3 and MegaDescriptor-L.  This measures species transfer relative to task-specific Re-ID supervision. Note that all of these models were pretrained on ImageNet data~\cite{deng2009imagenet}, where the classes 'tusker', 'African elephant', and 'Indian elephant' are present. 
Retrieval remains closed-set because every evaluated query identity occurs in
the gallery.

\subsection{Evaluation Protocols}
\label{sec:experiments:protocols}
Closed-set retrieval and AnimalCLEF-style association and
discovery use separate splits and evaluation procedures~\cite{animalclef2025,animal-clef-2026,elbassat2026calibrated}.

\paragraph{Closed-set retrieval.}
We construct observation groups using encounter metadata or timestamps,
exact-image matches, and within-identity perceptual similarity. We group
same-identity images when their dHash distance is $\leq4$ and their pHash
distance is $\leq6$. Cross-identity perceptual matches are retained and recorded for audit as they represent difficult cases. This
grouping is distinct from the corpus duplicate-removal step in
Sec.~\ref{sec:dataset:curation}.
Identities with fewer than two groups are excluded from retrieval. For an
identity with $G$ groups, we assign approximately 80\% to the gallery, clipping
the rounded count to $[1,G-1]$. Timestamps define an initial grouping and then a deterministic shuffle keyed by dataset,
identity, and seed assigns whole groups to the gallery. 

We fix splits for seeds $42$, $123$, $3407$, $2026$, and $1337$ and reuse the
same manifests across models. Group-disjointness checks ensure that recorded
observations and detected within-identity duplicate groups do not cross roles.
We rank gallery images by cosine similarity between $\ell_2$-normalized
embeddings and assign the identity of the highest-ranked image.

\paragraph{Dataset-adapted open-set association and discovery.}
We use the AnimalCLEF 2026 calibrated-similarity and graph-clustering
pipeline~\cite{animal-clef-2026,elbassat2026calibrated}. We assign
observation-aware roles using exact-image matches, dates, encounter identifiers,
and naming conventions. When temporal or encounter structure is insufficient,
we group images by DINOv2 embedding similarity~\cite{oquab2023dinov2}. Approximately 20\% of eligible identities,
along with identities that have only one observation group, enter
\texttt{query\_novel}. The remaining identities provide observation-disjoint
\texttt{gallery} and \texttt{query\_known} samples. When temporal structure is
available, later groups serve as queries. Private jaguar data are evaluated
without a gallery to test the quality of the clustering output when evaluating on unknown identities.
Calibration and parameter selection are described in Supplementary
Sec.~\ref{sec:suppl:per_dataset_open_set}.




\subsection{Evaluation Metrics and Statistical Analysis}
\label{sec:experiments:metrics}

\paragraph{Closed-set retrieval.}
We report Rank-1, CMC through rank 50, mAP, mINP~\cite{ye2021deep},
identity-balanced mAP and identity-balanced Rank-1. We rank gallery images individually and
consider all images of the query identity relevant. Identity-balanced metrics
first average scores within each query identity and then across identities.
For each dataset, we average query-mean metrics across five splits, then report
the equally weighted mean across the four datasets. Supplementary
Sec.~\ref{sec:suppl:metric_details} provides full definitions and secondary
analyses.

\paragraph{Open-set association and discovery.}
We report ARI~\cite{hubert1985comparing}, AMI~\cite{JMLR:v11:vinh10a},
V-measure~\cite{rosenberg2007v}, item-averaged B$^3$ precision, recall, and
F1~\cite{amigo2009comparison}, and pairwise clustering F1. B$^3$ F1 is the
harmonic mean of item-averaged precision and recall. Pairwise F1 treats pairs in the same predicted cluster as positive predictions
and pairs with the same true identity as positive ground-truth pairs. We compute
clustering metrics for all queries and for the novel-query subset. Gallery settings
also report known-attachment accuracy, identity-balanced correct known
recognition (BAKS), and identity-balanced novel rejection (BAUS), with
$\mathrm{NormAcc}=\sqrt{\mathrm{BAKS}\,\mathrm{BAUS}}$ computed per dataset.

Table~\ref{tab:open_world_macro} summarizes dataset-macro open-set association and discovery performance across the evaluated embedding and loss configurations. All-query ARI, B$^3$ F1, and pairwise F1 are averaged across all four datasets, whereas novel-only ARI, BAKS, and NormAcc are averaged across the three gallery-based datasets (Bristol, Elephants, and GCN-ID). NormAcc is computed per dataset as the geometric mean of BAKS and BAUS before macro-averaging. The corresponding per-dataset results, including gallery-free Jaguar discovery, are reported in Supplementary Table~\ref{tab:open_world_results}.
\paragraph{Paired statistical analysis.}
For objectives evaluated on the same backbone, we compute pointwise
95\% percentile intervals for paired metric differences using 10,000
identity-block bootstrap replicates. Identities are resampled with
replacement independently within each dataset, and each sampled identity
carries all its queries across the five splits and both compared models.
Each replicate preserves the split-then-dataset averaging procedure
described above. Intervals are pointwise and not adjusted for multiple
comparisons.

\FloatBarrier
\section{Results and Analysis}

While evaluating the performance of different models on closed-set and open-set re-identification, we noticed that models and losses that performed strongly on closed-set re-identification do not necessarily exhibit the same behavior on open-set re-identification.

\label{sec:results}

\subsection{Closed-Set Retrieval}
\label{sec:results:retrieval}

Table~\ref{tab:closed_set_results} reports dataset-macro retrieval;
Supplementary Table~\ref{tab:suppl:closed_set_per_dataset} provides the
per-dataset breakdown.

\begin{table}[!htbp]
\centering
\caption{Dataset-macro closed-set retrieval over four external datasets and
five split seeds (percentages). ID-balanced metrics average queries within each
identity before averaging identities. Bold marks the best value within each
backbone family, including ties at the reported precision. Pretrained rows
use the original animal Re-ID checkpoints without ApexID fine-tuning under
the same evaluation protocol.}
\label{tab:closed_set_results}
\footnotesize
\setlength{\tabcolsep}{2.2pt}
\begin{tabular}{@{}l r r r r r@{}}
\toprule
\textbf{Loss} & \textbf{R-1} & \textbf{mAP} &
\textbf{mINP} & \textbf{\shortstack{ID-bal.\\mAP}} &
\textbf{\shortstack{ID-bal.\\R-1}} \\
\midrule
\multicolumn{6}{l}{\textit{ConvNeXt-Small}} \\
ArcFace & 62.24 & 29.54 & 14.57 & 30.08 & 59.36 \\
Dynamic ArcFace & 61.75 & 29.23 & 14.51 & 29.62 & 58.72 \\
SphereFace2 & 63.60 & 30.24 & 15.31 & 30.83 & 60.78 \\
ApexID Hierarchical & \textbf{64.89} & \textbf{30.84} &
\textbf{15.71} & \textbf{31.72} & \textbf{62.13} \\
\midrule
\multicolumn{6}{l}{\textit{MIEW-ID v3}} \\
ArcFace & 63.65 & 32.09 & 17.71 & 32.60 & 60.82 \\
Dynamic ArcFace & 62.72 & 31.51 & 17.31 & 32.06 & 60.07 \\
SphereFace2 & 63.40 & 31.89 & 17.78 & 32.44 & 60.64 \\
ApexID Hierarchical & \textbf{63.88} & \textbf{32.54} &
\textbf{18.28} & \textbf{33.23} & \textbf{61.62} \\
Pretrained & 58.14 & 27.40 & 13.74 & 27.74 & 55.54 \\
\midrule
\multicolumn{6}{l}{\textit{MegaDescriptor-L}} \\
ArcFace & 58.20 & 27.13 & 13.05 & 27.53 & 55.52 \\
Dynamic ArcFace & 55.44 & 25.51 & 11.75 & 25.49 & 52.34 \\
SphereFace2 & 57.13 & 26.61 & 12.47 & 26.79 & 54.43 \\
ApexID Hierarchical & \textbf{61.18} & \textbf{29.07} &
\textbf{14.69} & \textbf{29.85} & \textbf{58.52} \\
Pretrained & 54.01 & 23.62 & 10.66 & 24.11 & 50.82 \\
\bottomrule
\end{tabular}
\end{table}

Within every backbone family, ApexID Hierarchical Loss is best on all five
reported macro retrieval metrics. With MIEW-ID v3, its $32.54\%$ mAP and
$18.28\%$ mINP are the highest of all fourteen configurations, and
ConvNeXt-Small with ApexID Hierarchical Loss attains the highest Rank-1
overall ($64.89\%$).
All eight fine-tuned animal Re-ID configurations improve dataset-macro
Rank-1 and mAP over their corresponding original checkpoints.

\paragraph{Paired objective comparisons.}
Figure~\ref{fig:suppl:paired} summarizes comparisons of ApexID Hierarchical Loss with each
baseline on the same backbone and frozen query records.
All nine macro-mAP differences are positive, ranging from $+0.45$
(MIEW-ID v3 vs.\ ArcFace) to $+3.56$ percentage points (MegaDescriptor-L
vs.\ Dynamic ArcFace), and none of their pointwise 95\% bootstrap intervals
includes zero. The gains are largest on MegaDescriptor-L ($+1.94$ to $+3.56$).

\paragraph{Per-dataset behavior.}
Against SphereFace2, the largest mAP gains occur on GCN-ID for all three
backbones ($+1.00$, $+1.27$, and $+7.57$ percentage points for ConvNeXt-S,
MIEW-ID v3, and MegaDescriptor-L); for MIEW-ID v3, the gain on the private
jaguar data is equally large. mAP on the private jaguar data increases for
every backbone (Supplementary Sec.~\ref{sec:suppl:closed_set}).

\paragraph{Transfer to unseen species.}
On ELPephants, ConvNeXt-Small with ApexID Hierarchical Loss achieves $17.69\%$ Rank-1
and $6.63\%$ mAP, the highest among all fourteen configurations. The
original MIEW-ID v3 checkpoint reaches $14.12\%$ Rank-1 and $4.82\%$ mAP,
while the original MegaDescriptor-L reaches $8.63\%$ and $3.34\%$,
respectively. A full comparison is shown in Fig.~\ref{fig:suppl:elephant} of the appendix.

\paragraph{Backbone capacity.}
The CNN-based ConvNeXt-S and MIEW-ID v3 backbones have 49.45M and 51.08M
parameters, respectively, versus 195.20M for MegaDescriptor-L (Supplementary
Table~\ref{tab:suppl:backbone_params}). Despite this
roughly fourfold size gap, the family-wise best ConvNeXt-S and MIEW-ID v3 values
exceed the best MegaDescriptor-L values on all five macro retrieval metrics.

\subsection{Representation and Verification}
\label{sec:results:representation}
Against SphereFace2, ApexID Hierarchical Loss increases same-category AUROC in 9 of 12 backbone--dataset combinations and identity silhouette scores in 8 of 12 (Supplementary Table~\ref{tab:suppl:representation}). Fisher ratios improve in all 12 combinations, while low-FAR verification includes regressions. These embedding-space diagnostics capture complementary properties: AUROC measures how well same-category pairs are separated from different-category pairs~\cite{fawcett2006introduction}, silhouette scores characterize within-identity cohesion relative to between-identity separation~\cite{rousseeuw1987silhouettes}, Fisher ratios quantify between-identity dispersion relative to within-identity dispersion, and effective rank summarizes the dimensionality of the embedding distribution~\cite{roy2007effective}.
\subsection{Dataset-Adapted Open-Set Analysis}
\label{sec:results:openworld}

Table~\ref{tab:open_world_macro} reports macro association and discovery
scores for fourteen embeddings under the dataset-adapted protocol.
Supplementary Table~\ref{tab:open_world_results} gives per-dataset results,
including gallery-free jaguar discovery.

\begin{table}[!htbp]
\centering
\caption{Dataset-macro open-set association and discovery results (split seed~42).
All-query ARI, B$^3$ F1, and pairwise F1 are averaged over four datasets.
Novel-only ARI, BAKS, and Normalized Accuracy (NormAcc) are averaged over the three gallery
datasets (Bristol, elephants, and GCN-ID). Boldface marks the best value
within each backbone family.}
\label{tab:open_world_macro}
\footnotesize
\setlength{\tabcolsep}{2.2pt}
\begin{tabular}{@{}l r r r r r r@{}}
\toprule
\textbf{Loss} & \textbf{ARI} & \textbf{B$^3$ F1} &
\textbf{\shortstack{Pair\\F1}} & \textbf{\shortstack{Nov.\\ARI}} & \textbf{BAKS} & \textbf{NormAcc} \\
\midrule
\multicolumn{7}{l}{\textit{ConvNeXt-S}} \\
ArcFace & 0.424 & 0.596 & 0.447 & 0.392 & 0.098 & 0.207 \\
Dynamic ArcFace & 0.436 & \textbf{0.597} & 0.460 & 0.397 & 0.305 & \textbf{0.338} \\
SphereFace2 & \textbf{0.441} & 0.588 & \textbf{0.463} & \textbf{0.404} & 0.305 & \textbf{0.338} \\
ApexID Hierarchical & 0.432 & 0.591 & 0.457 & 0.403 & \textbf{0.319} & 0.322 \\
\midrule
\multicolumn{7}{l}{\textit{MIEW-ID v3}} \\
ArcFace & 0.433 & 0.582 & 0.457 & 0.409 & 0.320 & 0.400 \\
Dynamic ArcFace & 0.444 & 0.593 & \textbf{0.479} & 0.396 & 0.331 & 0.419 \\
SphereFace2 & 0.427 & \textbf{0.594} & 0.452 & 0.400 & \textbf{0.431} & \textbf{0.537} \\
ApexID Hierarchical & 0.430 & 0.593 & 0.465 & 0.406 & 0.379 & 0.437 \\
Pretrained & \textbf{0.451} & 0.592 & 0.474 & \textbf{0.416} & 0.405 & 0.521 \\
\midrule
\multicolumn{7}{l}{\textit{MegaDescriptor-L}} \\
ArcFace & 0.414 & 0.595 & 0.437 & 0.384 & 0.419 & \textbf{0.501} \\
Dynamic ArcFace & \textbf{0.443} & \textbf{0.609} & \textbf{0.473} & 0.382 & \textbf{0.475} & 0.498 \\
SphereFace2 & 0.411 & 0.567 & 0.427 & 0.383 & 0.321 & 0.346 \\
ApexID Hierarchical & 0.434 & 0.601 & 0.463 & 0.384 & 0.307 & 0.316 \\
Pretrained & 0.428 & 0.580 & 0.457 & \textbf{0.398} & 0.317 & 0.321 \\
\bottomrule
\end{tabular}
\end{table}

\textbf{High scores on closed-set re-identification do not imply high scores on open-set re-identification}. Macro
ARI spans only $0.411$--$0.451$, as GCN-ID is nearly saturated and the other
datasets are difficult for every model, as shown in Supplementary Table~\ref{tab:open_world_results}.
Within ConvNeXt-S, ApexID Hierarchical Loss gives the best BAKS, while SphereFace2
gives the best all-query ARI, pairwise F1, and novel-only ARI, and ties with
Dynamic ArcFace for the best Normalized Accuracy. Within MIEW-ID v3, the unmodified checkpoint leads in all-query ARI, while SphereFace2
leads in known-individual attachment. Within MegaDescriptor-L, ApexID Hierarchical Loss ranks second to Dynamic
ArcFace on ARI, B$^3$ F1, and pairwise F1, but has the lowest BAKS and
Normalized Accuracy in its family. ApexID Hierarchical Loss has the highest
elephant ARI among the fine-tuned objectives for ConvNeXt-S and
MegaDescriptor-L, but the lowest ARI on private jaguar data for ConvNeXt-S
and MIEW-ID v3. On elephants, the unmodified MIEW-ID v3 checkpoint has the highest ARI
overall ($0.285$), while known-individual recognition remains near zero for
every embedding (BAKS $\leq0.014$).

\subsection{Ablation: Category Loss Weight}
\label{sec:results:ablation}

To isolate the contribution of category supervision, we vary the category
loss weight $\lambda_{\mathrm{SP}}$ in Eq.~\eqref{eq:scsf2:total} over
$\{0.025, 0.05, 0.075, 0.10, 0.15, 0.25, 0.50\}$ on ConvNeXt-Small. All other
settings are held fixed. The SphereFace2 baseline from
Table~\ref{tab:closed_set_results} serves as the reference without category
supervision. Table~\ref{tab:ablation_category_weight} reports dataset-macro
closed-set retrieval under the protocol of Sec.~\ref{sec:experiments:protocols}.

All main results use $\lambda_{\mathrm{SP}}=0.50$, the best setting on four of the five metrics.
\begin{table}[!htbp]
\centering
\caption{Effect of the category loss weight $\lambda_{\mathrm{SP}}$ on
dataset-macro closed-set retrieval with ConvNeXt-Small (percentages), using the
same datasets, split seeds, and metrics as Table~\ref{tab:closed_set_results}.
SphereFace2 is trained without category supervision.}
\label{tab:ablation_category_weight}
\footnotesize
\setlength{\tabcolsep}{2.2pt}
\begin{tabular}{@{}l r r r r r@{}}
\toprule
\textbf{Objective} & \textbf{R-1} & \textbf{mAP} &
\textbf{mINP} & \textbf{\shortstack{ID-bal.\\mAP}} & \textbf{\shortstack{ID-bal.\\R-1}} \\
\midrule
SphereFace2 (no category term) & 63.603 & 30.244 & 15.310 & 30.825 & 60.779 \\
\midrule
$\lambda_{\mathrm{SP}}=0.025$ & 64.257 & 30.499 & 15.334 & 31.113 & 61.260 \\
$\lambda_{\mathrm{SP}}=0.05$  & 64.262 & 30.586 & 15.375 & 31.334 & 61.382 \\
$\lambda_{\mathrm{SP}}=0.075$ & 64.193 & 30.542 & 15.383 & 31.114 & 61.047 \\
$\lambda_{\mathrm{SP}}=0.10$  & 64.602 & 30.662 & 15.531 & 31.235 & 61.409 \\
$\lambda_{\mathrm{SP}}=0.15$  & 64.478 & 30.546 & 15.333 & 31.257 & 61.730 \\
$\lambda_{\mathrm{SP}}=0.25$  & 64.796 & 30.724 & 15.493 & 31.510 & \textbf{62.303} \\
$\lambda_{\mathrm{SP}}=0.50$  & \textbf{64.888}& \textbf{30.835} & \textbf{15.715} & \textbf{31.724} & 62.125\\
\bottomrule
\end{tabular}
\end{table}

\FloatBarrier
\section{Limitations}
\label{sec:limitations}

ApexID assumes a two-level category--identity hierarchy, whose effectiveness may depend on the granularity and reliability of the animal category annotations. These categories represent broad animal types rather than a consistent biological taxonomic level. In particular, they do not necessarily distinguish between species within the same broader animal category. For example, all individuals in the whale subset are assigned to a single ``whale'' category, without distinguishing among whale species. This choice reflects the available annotations and uncertainty regarding species-level identification.


\section{Conclusion}
\label{sec:conclusion}

We presented the ApexID Hierarchical Loss, a multi-species animal Re-ID
loss function that extends SphereFace2 by combining global identity
discrimination with balanced category supervision. Observation-aware evaluation across three model families shows that ApexID
improves closed-set retrieval over the original pretrained references, while
its gains in open-set re-identification vary by backbone and dataset. Open-set
analyses reveal complementary strengths and remaining discovery
challenges. These findings support category-aware representations for
wildlife re-identification and underscore the need to validate transfer across
species and acquisition conditions.
\label{page:main_end}
\section*{LLM Usage Statement}

We used LLMs for language editing, grammar correction, figure editing and design, and improving clarity throughout the manuscript.
The authors reviewed and revised all AI-assisted text and take full
responsibility for the paper's content.

\FloatBarrier
{
    \small
    \bibliographystyle{ieeenat_fullname}
    \bibliography{main}
}


\clearpage
\setcounter{page}{1}
\maketitlesupplementary
\raggedbottom

\makeatletter
\setlength{\@dblfptop}{0pt}
\setlength{\@dblfpsep}{10pt}
\setlength{\@dblfpbot}{0pt plus 1fil}
\setlength{\@fpsep}{10pt}
\makeatother
\renewcommand{\dbltopfraction}{0.95}
\setcounter{dbltopnumber}{4}
\renewcommand{\textfraction}{0.05}
\renewcommand{\dblfloatpagefraction}{0.8}
\setcounter{section}{0}
\renewcommand{\thesection}{\Alph{section}}
\renewcommand{\theHsection}{supplementary.\arabic{section}}

References use the main paper's bibliography; table and figure numbering is
shared across the main paper and supplementary material.

\section{Data Sources and Curation Details}
\label{sec:suppl:data}
Table~\ref{tab:apexid_train_datasets} expands the corpus summary in
Sec.~\ref{sec:dataset:sources}. Identity labels are source-specific; image
and observation totals include both training and validation partitions.
\begin{table*}[t]
\centering
\caption{ApexID corpus after preprocessing and curation. Train IDs count
source-specific identities with at least one training image. Image and
observation totals cover both partitions; final columns give train and validation
image counts. Observation groups may contain multiple images.
Train IDs exclude the 1,671 held-out-identity validation identities, which
have no training images; adding them gives the 13,348 unique identities in
Table~\ref{tab:apexid_corpus_summary}.}
\label{tab:apexid_train_datasets}
\resizebox{\textwidth}{!}{%
\begin{tabular}{l l r r r r r}
\toprule
\textbf{Dataset} & \textbf{Animal Category} & \textbf{Train IDs} &
\textbf{Images} & \textbf{Observations} & \textbf{Train images} &
\textbf{Val. images} \\
\midrule
AmvrakikosTurtles~\cite{adam2025exploiting} & Sea turtle & 41 & 141 & 100 & 59 & 82 \\
BalearicLizard~\cite{alcaraz2026long} & Lizard & 656 & 4,359 & 4,330 & 2,859 & 1,500 \\
CTai~\cite{freytag2016chimpanzee} & Chimpanzee & 57 & 3,609 & 3,608 & 2,443 & 1,166 \\
CzechLynx~\cite{picek2026czechlynx} & Lynx & 279 & 37,829 & 10,971 & 28,792 & 9,037 \\
GiraffeZebraID~\cite{parham2017animal} & Giraffe; zebra & 876 & 4,333 & 1,455 & 3,569 & 764 \\
Giraffes~\cite{miele2021revisiting} & Giraffe & 136 & 892 & 323 & 651 & 241 \\
HappyWhale~\cite{cheeseman2026happywhale} & Whales; dolphins & 5,538 & 41,407 & 41,318 & 26,304 & 15,103 \\
HyenaID2022~\cite{botswana2022hyena} & Hyena & 211 & 2,422 & 2,309 & 1,599 & 823 \\
IPanda50~\cite{wang2021giant} & Panda & 42 & 5,167 & 1,457 & 3,619 & 1,548 \\
Jaguar\_data~\cite{rueda2025jaguars} & Jaguar & 27 & 1,178 & 1,000 & 795 & 383 \\
LeopardID2022~\cite{botswana2022leopard} & Leopard & 251 & 4,953 & 4,269 & 3,494 & 1,459 \\
NDD20~\cite{trotter2020ndd20} & Dolphin & 68 & 2,122 & 2,117 & 1,488 & 634 \\
NyalaData~\cite{vcermak2024wildlifedatasets} & Nyala & 150 & 1,442 & 867 & 951 & 491 \\
ReunionTurtles~\cite{adam2025exploiting} & Sea turtle & 69 & 237 & 169 & 104 & 133 \\
SalamanderID2025~\cite{animalclef2025} & Salamander & 242 & 1,078 & 1,019 & 675 & 403 \\
SealID~\cite{nepovinnykh2022sealID} & Seal & 48 & 1,668 & 1,662 & 1,101 & 567 \\
WhaleSharkID~\cite{holmberg2009estimating} & Whale shark & 388 & 5,958 & 5,759 & 3,855 & 2,103 \\
ZakynthosTurtles~\cite{adam2025exploiting} & Sea turtle & 33 & 114 & 81 & 47 & 67 \\
HumpbackWhaleID~\cite{vcermak2024wildlifedatasets} & Whale & 2,565 & 13,624 & 13,621 & 8,516 & 5,108 \\
\midrule
\textbf{Total} & \textbf{Multi-species} & \textbf{11,677} &
\textbf{132,533} & \textbf{96,435} & \textbf{90,921} &
\textbf{41,612} \\
\bottomrule
\end{tabular}%
}

\end{table*}

\paragraph{Curation sequence.}
When acquisition information is available we use it to separate test and training set images. 
Within-identity MegaDescriptor similarity is used to remove near-duplicate images before
segmentation and partitioning.
SAM~3 is prompted with an animal-category description, such as ``photo of a giraffe''; the
highest-confidence valid mask is then applied with a black background. V-Rush
review checks image quality, masks, and annotations; MobileSAM bounding-box
prompts and manual brush correction are used to refine incomplete masks.


\subsection{Training and Validation Partitioning}

\label{sec:suppl:partitioning}

An \emph{identity} denotes an annotated individual animal. An
\emph{observation} groups images associated with an encounter, capture event,
or acquisition sequence. Images from the same observation may share pose,
illumination, background, and camera conditions, making image-level random
partitioning susceptible to correlated samples across splits.

We write \texttt{val\_known} for the known-identity validation partition
and \texttt{val\_unknown} for the held-out-identity validation partition of
Table~\ref{tab:apexid_corpus_summary}.
Following time- and similarity-aware animal Re-ID
evaluation~\cite{Adam_2025_CVPR}, known-identity data are partitioned at the
\textbf{observation-group} level. Fully dated identities spanning multiple
calendar days use a chronological suffix closest to a 22.5\% validation-image
target while keeping earlier dates in training. For identities covered by
verified filename or session parsers (CTai, Giraffes, IPanda50, NDD20, and
NyalaData), a deterministic subset of whole groups is selected to approach the
same target. For all remaining identities, frozen DINOv2
features~\cite{oquab2023dinov2} are used to form within-identity connected
components, with exact-duplicate links identified first. Multi-image
components are assigned entirely to training;  singleton (single-image)
components may enter \texttt{val\_known}. Dataset-wise selection at a 12.5\% ratio routes complete
unknown identities to \texttt{val\_unknown} without splitting their groups. An
identity may therefore appear in both training and \texttt{val\_known} through
\emph{different} observation groups, but images sharing a group always share a
partition.
Because whole identities and whole observation groups are assigned
together, small datasets with few groups per identity can exceed the nominal
targets. AmvrakikosTurtles, ReunionTurtles, and ZakynthosTurtles therefore
have more validation than training images
(Table~\ref{tab:apexid_train_datasets}).

\paragraph{External subset scope.}
Table~\ref{tab:apexid_test_datasets} describes subsets eligible for closed-set
evaluation.
Closed-set eligibility excludes identities with fewer than two observation
groups. The separate open-set protocol may include such identities as novel
queries and uses a gallery-free setting for private jaguar data. Role
construction for this protocol is specified in
Sec.~\ref{sec:experiments:protocols}.
\section{Implementation and Metric Details}
\label{sec:suppl:implementation}

\paragraph{Preprocessing and embedding extraction.}
All models use the same segmented source images, with checkpoint-specific
resizing and normalization. The ConvNeXt-Small preset uses RGB foreground
images resized to $440\times 440$ and normalized with ImageNet channel
statistics. MIEW-ID v3 uses the same preprocessing; MegaDescriptor-L uses
$384\times384$ inputs normalized with mean and standard deviation $0.5$.
ConvNeXt-Small adds a learned linear projection to a 2048-dimensional
embedding, whereas MIEW-ID v3 (2152) and MegaDescriptor-L (1536) use their
native embeddings. We keep the ConvNeXt-Small projection, discard all training
classification heads, and
$\ell_2$-normalize the extracted embeddings before evaluation.

Frozen manifests and cached embeddings separate data preparation, feature
extraction, and evaluation. This allows the same prepared samples and
extracted representations to be reused across the relevant evaluation
procedures without repeating inference.

\paragraph{ConvNeXt-Small training preset.}
Batches use identity-uniform $P \times K_{\mathrm{PK}}$ sampling, with
$P=16$ identities and $K_{\mathrm{PK}}=4$ images per identity
($B=64$), following \cite{hermans2017defense,schroff2015facenet}.
When an identity has fewer than four available images, its images are
sampled with replacement. This affects batch composition only; proxy
objectives consider all training identities.

Augmentation includes mild affine transforms and occasional grayscale
conversion, blur, Gaussian noise, and pixel dropout. Validation uses
deterministic resizing and normalization.

\paragraph{Optimization.} We fine-tune all parameters with AdamW (weight decay $10^{-4}$), using learning rates of $10^{-5}$ for the embedding network, including the ConvNeXt-Small projection, and $10^{-4}$ for the loss parameters (identity and category proxies and the bias). The learning rate warms up linearly from $10^{-6}$ over five epochs and then follows cosine annealing to $10^{-6}$. The shared bias $b$ in Eq.~\eqref{eq:scsf2:identity_logit} uses the SphereFace2 initialization~\cite{wen2022sphereface2}. All models train for 150 epochs with gradient-norm clipping at 5.0.

\paragraph{ApexID Hierarchical Loss preset.}
The released ApexID Hierarchical Loss preset uses
$K_{\mathrm{SC}}=2$ SubCenter proxies per identity,
$m_{\mathrm{ID}}=0.30$, $m_{\mathrm{SP}}=0.10$ radians,
$r=\gamma=40$, $t=3$, $\alpha=0.9$, $\eta=20$,
and $\lambda_{\mathrm{SP}}=0.50$ for all three backbones. MIEW-ID v3 and
ConvNeXt-Small use $440\times440$ inputs and MegaDescriptor-L uses
$384\times384$.
Here, $K_{\mathrm{SC}}$ denotes the number of SubCenter proxies
maintained for each identity, whereas $K_{\mathrm{PK}}$ denotes the
number of training images sampled per identity in each PK batch.

\subsection{Metric Definitions and Secondary Analyses}
\label{sec:suppl:metric_details}
\paragraph{Image retrieval.}
We report Rank-1, cumulative matching characteristic (CMC) curves up to
rank 50, mean average precision (mAP), mean inverse negative penalty
(mINP)~\cite{ye2021deep}, and identity-balanced Rank-1 and mAP.
Rank-$k$, equivalently CMC@$k$, is the fraction of queries with at least
one correct match among the first $k$ ranked gallery images. Multiple
images of one gallery identity occupy separate ranks. Average precision
is the mean precision at the ranks of all relevant gallery images, where
relevance denotes a shared identity; mAP averages query-level average
precision. For each query, inverse negative penalty is the number of
relevant gallery images divided by the rank of the last relevant image;
mINP is its mean over queries. Identity-balanced metrics first average
within each query identity and then equally across identities.

\paragraph{Pairwise clustering F1.}
Consider unordered pairs of distinct query images. A predicted-positive pair
shares a predicted label; a ground-truth-positive pair shares a true identity.
True positives satisfy both conditions. Pairwise precision and recall are
$P_{\mathrm{pair}}=\mathrm{TP}/(\mathrm{TP}+\mathrm{FP})$ and
$R_{\mathrm{pair}}=\mathrm{TP}/(\mathrm{TP}+\mathrm{FN})$; their harmonic mean
gives pairwise F1. This differs from the item-averaged B$^3$ metric.

\paragraph{Secondary analyses.}
We report same-category verification (AUROC, EER, and TAR at fixed FAR),
embedding diagnostics, and robustness to limited galleries and same-category
distractors. Same-category distractors are negative matches of individuals of the same species. Conditional retrieval excludes queries lacking an eligible
gallery positive and reports resulting query coverage. Metric-specific
sample counts and gallery restrictions accompany the additional results.

\FloatBarrier

\section{Additional Closed-Set Results}
\label{sec:suppl:closed_set}

This section expands the dataset-macro results of
Sec.~\ref{sec:results:retrieval}. All fourteen configurations use the same
frozen protocol, split manifests, and query records; the 9,861 primary
query--split records per model comprise 6,508 distinct query images, 5,425
recorded observations, and 537 identities. 

\paragraph{Backbone parameter counts.}
Table~\ref{tab:suppl:backbone_params} reports backbone-only parameter counts.

\begin{table}[!htb]
\centering
\caption{Backbone-only parameter counts (millions).}
\label{tab:suppl:backbone_params}
\scriptsize
\setlength{\tabcolsep}{4pt}
\begin{tabular}{@{}l r@{}}
\toprule
\textbf{Model} & \textbf{Backbone parameters (M)} \\
\midrule
ConvNeXt-S (ApexID) & 49.45 \\
MIEW-ID v3 & 51.08 \\
MegaDescriptor-L & 195.20 \\
\bottomrule
\end{tabular}
\end{table}

\begin{figure}[!t]
\centering
\includegraphics[width=\linewidth]{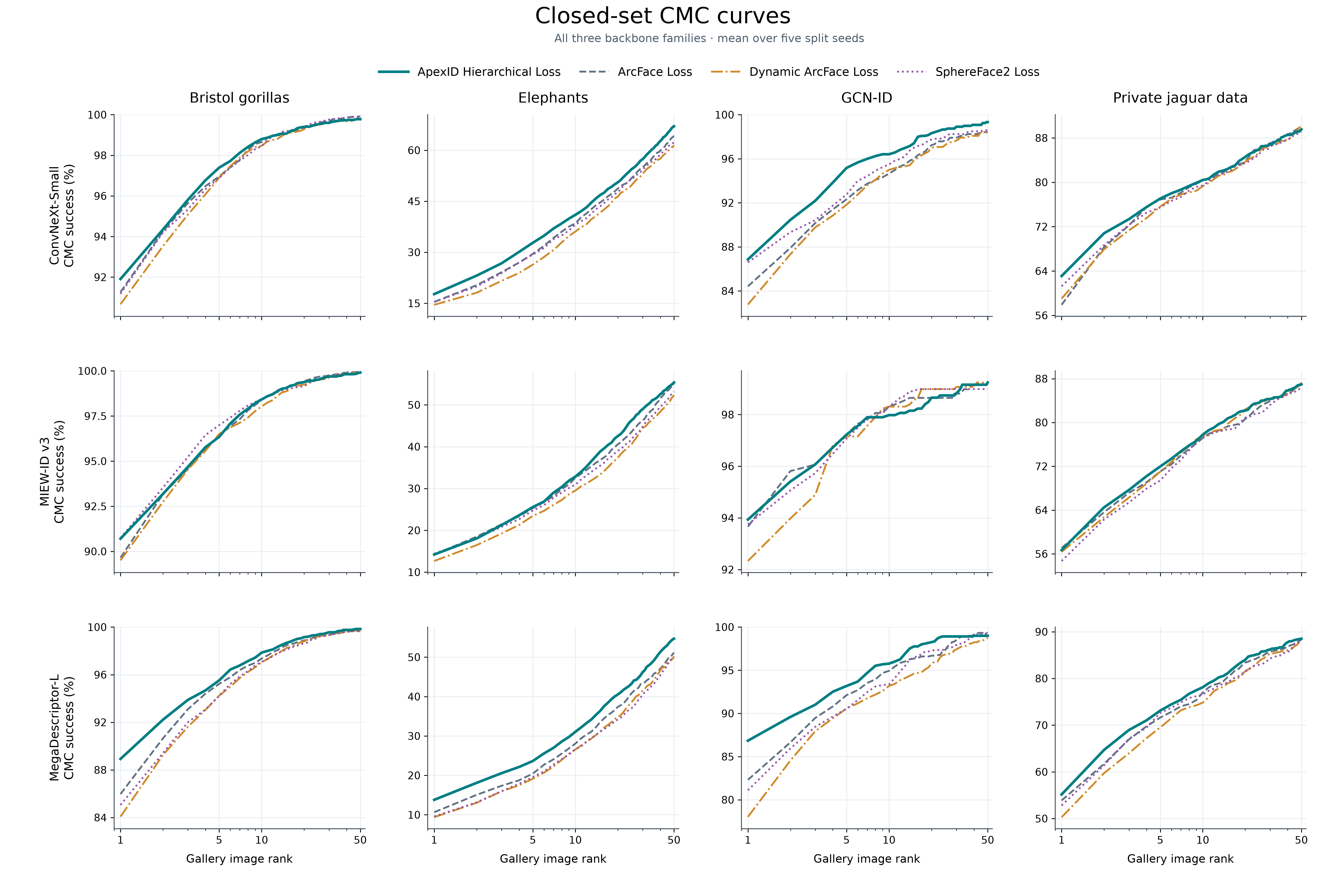}
\caption{CMC curves over ranks 1--50 for the twelve trained
checkpoints, averaged over the five split seeds.}
\label{fig:suppl:cmc}
\end{figure}

\paragraph{Per-dataset retrieval.}
Table~\ref{tab:suppl:closed_set_per_dataset} reports mAP, Rank-1, and mINP for
each dataset.
Different datasets favor different checkpoints: MIEW-ID ArcFace has the
highest mAP on Bristol, MIEW-ID ApexID Hierarchical on GCN-ID, and ConvNeXt-S
ApexID Hierarchical on elephants and the private jaguar data. High Rank-1 can
coexist with incomplete retrieval of all positives: MIEW-ID ApexID
Hierarchical reaches $90.71\%$ Bristol Rank-1 but only $26.48\%$ mAP and
$14.98\%$ mINP. Bristol contains only seven evaluated
identities, and each query has many gallery positives. Positive prevalence
and gallery sizes differ across datasets, so raw cross-dataset differences
are not controlled loss effects. CMC curves for the twelve trained
checkpoints are shown in Fig.~\ref{fig:suppl:cmc}.

\begin{table*}[!t]
\centering
\caption{Per-dataset closed-set retrieval (percentages; means over five split seeds),
including the two original pretrained references. Boldface marks the best
value within each backbone family, including ties at the reported precision.}
\label{tab:suppl:closed_set_per_dataset}
\scriptsize
\setlength{\tabcolsep}{2.5pt}
\begin{tabular}{@{}l l rrr rrr rrr rrr@{}}
\toprule
& & \multicolumn{3}{c}{\textbf{Bristol}} &
\multicolumn{3}{c}{\textbf{Elephants}} &
\multicolumn{3}{c}{\textbf{GCN-ID}} &
\multicolumn{3}{c}{\textbf{\shortstack{Private\\jaguar data}}} \\
\cmidrule(lr){3-5} \cmidrule(lr){6-8}
\cmidrule(lr){9-11} \cmidrule(lr){12-14}
\textbf{Backbone} & \textbf{Loss} &
mAP & R-1 & mINP &
mAP & R-1 & mINP &
mAP & R-1 & mINP &
mAP & R-1 & mINP \\
\midrule
ConvNeXt-S & ArcFace & \textbf{26.44} & 91.27 & \textbf{15.05} & 5.89 & 15.39 & 1.31 & 60.50 & 84.41 & 33.45 & 25.33 & 57.90 & 8.50 \\
ConvNeXt-S & Dynamic ArcFace & 25.84 & 90.67 & 15.03 & 5.32 & 14.54 & 1.07 & 60.35 & 82.76 & 33.53 & 25.41 & 59.01 & 8.38 \\
ConvNeXt-S & SphereFace2 & 26.06 & 91.18 & 15.02 & 5.72 & 15.38 & 1.18 & 62.96 & 86.60 & 36.37 & 26.23 & 61.25 & 8.67 \\
ConvNeXt-S & ApexID Hierarchical Loss & 26.30 & \textbf{91.91} & 15.01 & \textbf{6.63} & \textbf{17.69} & \textbf{1.49} & \textbf{63.96} & \textbf{86.86} & \textbf{37.16} & \textbf{26.46} & \textbf{63.10} & \textbf{9.20} \\
\midrule
MIEW-ID v3 & ArcFace & \textbf{26.95} & 89.65 & 14.99 & 4.74 & 14.21 & 0.80 & 73.85 & 93.66 & 48.06 & 22.81 & \textbf{57.07} & 6.98 \\
MIEW-ID v3 & Dynamic ArcFace & 26.09 & 89.50 & 14.98 & 4.29 & 12.66 & 0.76 & 72.98 & 92.33 & 46.50 & 22.66 & 56.37 & 7.00 \\
MIEW-ID v3 & SphereFace2 & 26.57 & \textbf{90.74} & 14.98 & 4.73 & \textbf{14.40} & 0.88 & 73.72 & 93.75 & 48.28 & 22.55 & 54.69 & 6.99 \\
MIEW-ID v3 & ApexID Hierarchical Loss & 26.48 & 90.71 & 14.98 & \textbf{4.87} & 14.21 & \textbf{0.92} & \textbf{74.99} & \textbf{93.93} & \textbf{49.68} & \textbf{23.82} & 56.68 & \textbf{7.53} \\
MIEW-ID v3 & Pretrained & 26.12 & 88.39 & \textbf{15.00} & 4.82 & 14.12 & 0.90 & 58.66 & 81.60 & 32.38 & 20.02 & 48.46 & 6.69 \\
\midrule
MegaDescriptor-L & ArcFace & 23.63 & 85.97 & 14.98 & 4.12 & 10.65 & 1.00 & 56.58 & 82.31 & 27.61 & 24.21 & 53.87 & 8.60 \\
MegaDescriptor-L & Dynamic ArcFace & 23.20 & 84.07 & 14.97 & 3.72 & 9.38 & 0.84 & 51.92 & 78.02 & 23.39 & 23.18 & 50.29 & 7.79 \\
MegaDescriptor-L & SphereFace2 & 23.73 & 85.05 & 14.98 & 3.87 & 9.57 & 0.99 & 54.60 & 81.10 & 25.90 & 24.25 & 52.80 & 8.03 \\
MegaDescriptor-L & ApexID Hierarchical Loss & \textbf{24.41} & \textbf{88.92} & \textbf{14.99} & \textbf{4.94} & \textbf{13.79} & \textbf{1.05} & \textbf{62.17} & \textbf{86.84} & \textbf{33.92} & \textbf{24.75} & \textbf{55.15} & \textbf{8.81} \\
MegaDescriptor-L & Pretrained & 23.26 & 84.02 & 14.97 & 3.34 & 8.63 & 0.77 & 47.46 & 73.29 & 18.92 & 20.40 & 50.09 & 8.00 \\
\bottomrule
\end{tabular}
\end{table*}

\begin{figure*}[!t]
\centering
\begin{minipage}[t]{0.485\textwidth}
\vspace{0pt}
\centering
\includegraphics[width=\linewidth]{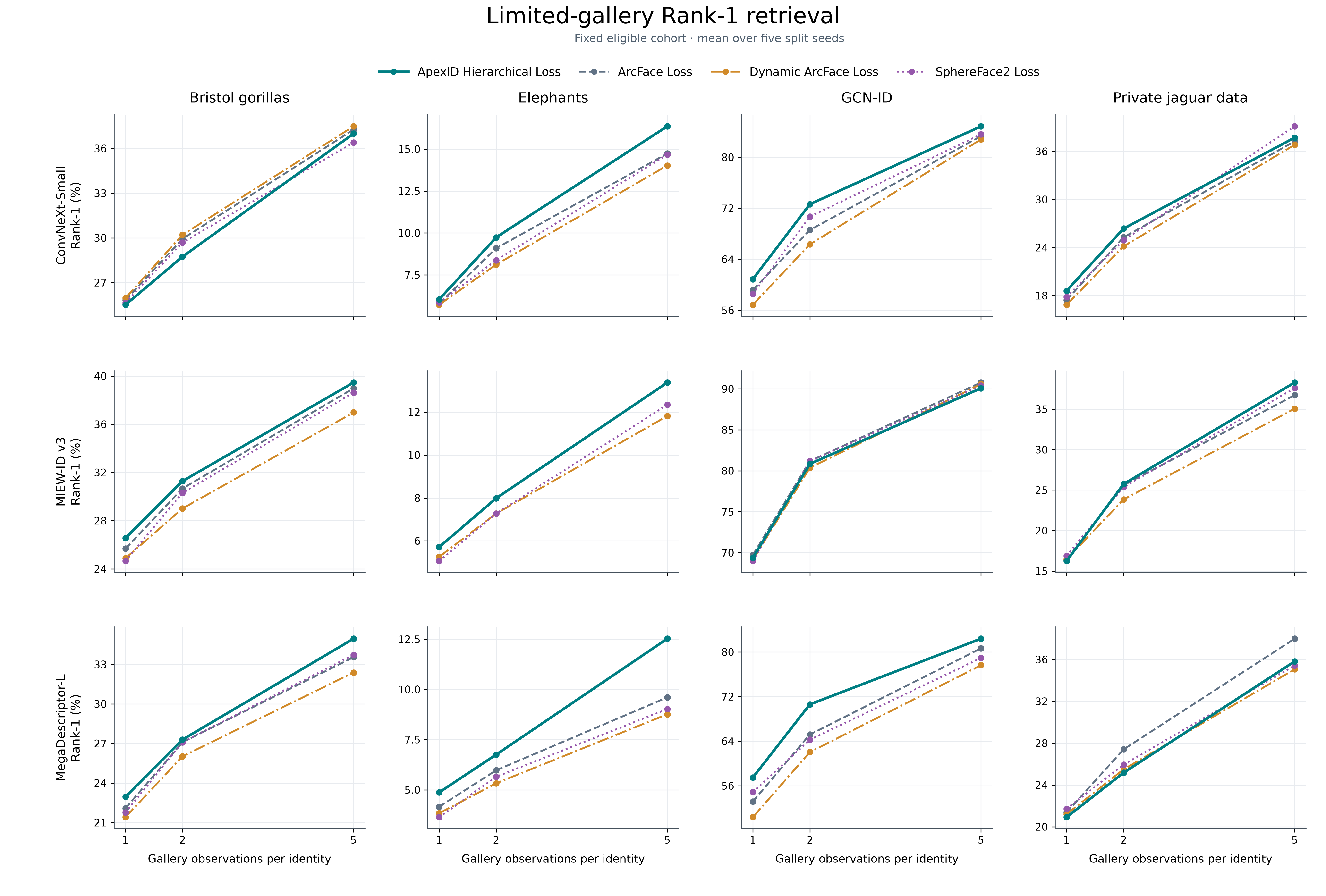}
\captionof{figure}{Rank-1 under limited gallery observations per identity,
with the same eligible cohort across $k$. Curves are five-split means.}
\label{fig:suppl:limited_gallery}
\end{minipage}\hfill
\begin{minipage}[t]{0.485\textwidth}
\vspace{0pt}
\centering
\includegraphics[width=\linewidth]{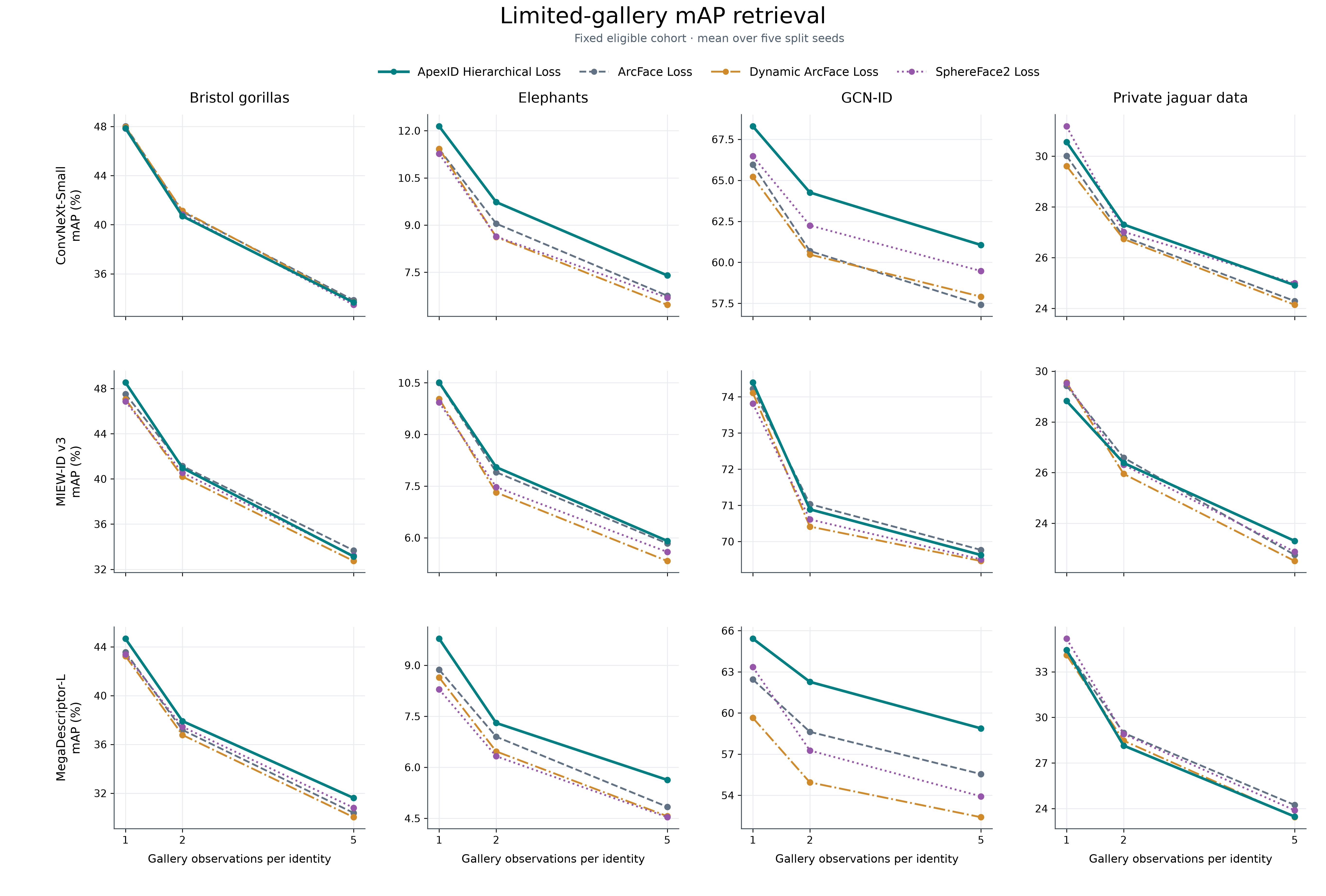}
\captionof{figure}{mAP under limited gallery observations per identity,
with the same eligible cohort across $k$. Curves are five-split means.}
\label{fig:suppl:limited_gallery_map}
\end{minipage}
\end{figure*}

\paragraph{Limited galleries.}
Fig.~\ref{fig:suppl:limited_gallery} restricts a fixed cohort of identities
to $k\in\{1,2,5\}$ gallery observations, using one image per recorded
observation. The cohort retains 7/7 Bristol gorillas (Bristol), 154/272 elephants, 73/207 great crested newt (GCN) identities,
and 26/51 identities in the private jaguar data. At
$k=5$, ConvNeXt-S ApexID Hierarchical is the best ConvNeXt-S objective on Bristol,
elephants, and GCN-ID, raising elephant Rank-1 from $14.67\%$ (SphereFace2) to
$16.55\%$, although it ties SphereFace2 at $5.84\%$ with a single observation at the
reported precision. MIEW-ID ApexID Hierarchical improves five-observation GCN-ID
Rank-1 from $90.39\%$ to $91.35\%$. The corresponding mAP curves are shown
in Fig.~\ref{fig:suppl:limited_gallery_map}; mAP can fall while Rank-1 rises
because both positive and negative gallery content change with $k$.

\begin{figure*}[!t]
\centering
\begin{minipage}[c]{0.58\textwidth}
\centering
\includegraphics[width=\linewidth]{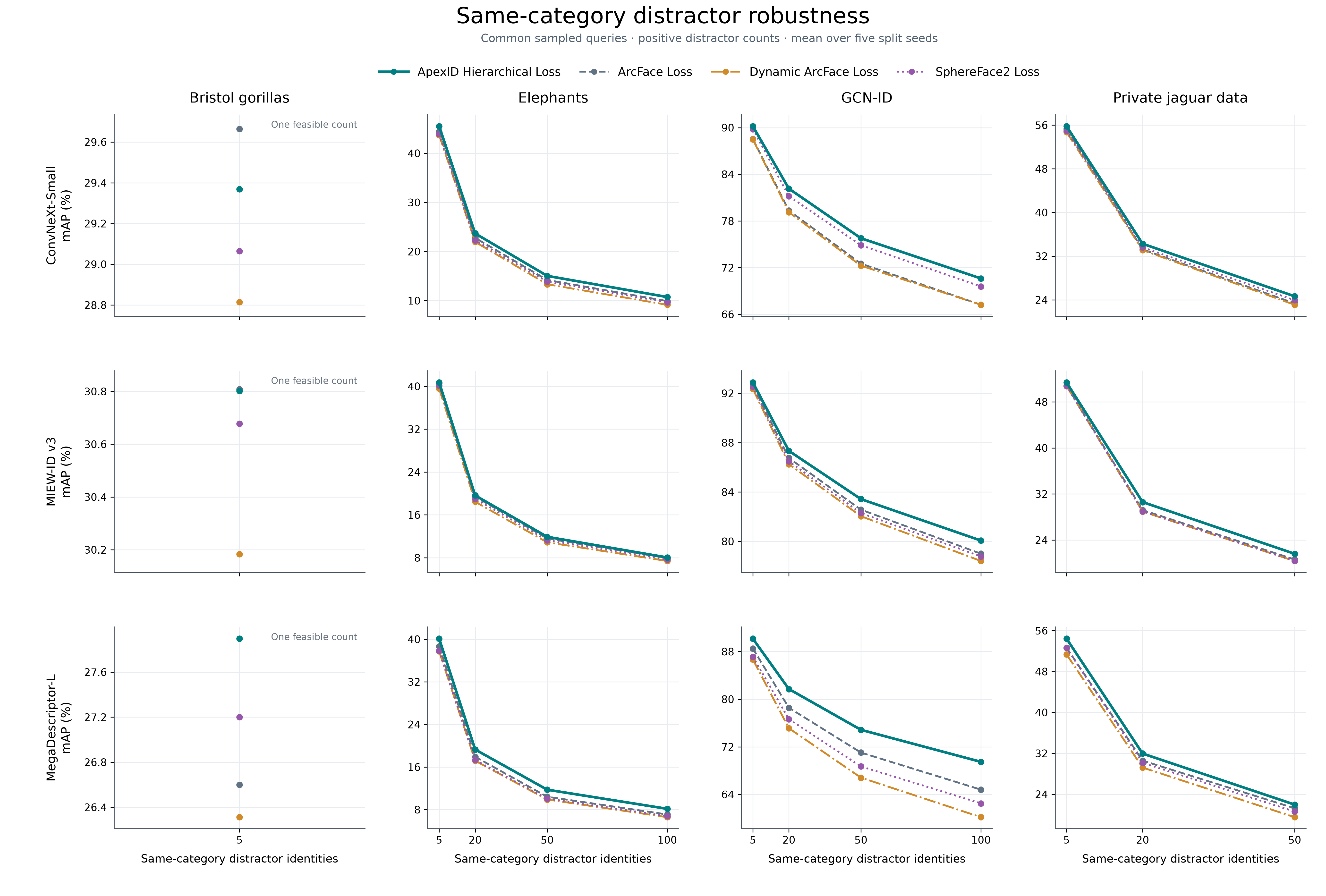}
\end{minipage}\hfill
\begin{minipage}[c]{0.39\textwidth}
\captionof{figure}{Mean average precision (mAP) as more individuals from the same category are added to the search gallery. For example, when the query shows one elephant, the distractors are other elephants. Within each dataset, the same query images are used at every distractor count. The Bristol Gorillas dataset has only seven distinct identities.}
\label{fig:suppl:distractors}
\end{minipage}
\end{figure*}

\paragraph{Same-category distractors.}
Fig.~\ref{fig:suppl:distractors} retains all positives and adds negative
identities from the same category, up to 5 for Bristol, 50 for the private
jaguar data, and 100 for elephants and GCN-ID; these endpoints are not
equally difficult. At 100 elephant distractors, ApexID Hierarchical retains higher
mAP than SphereFace2 on all three backbones ($10.62\%$ vs.\ $9.65\%$, $7.93\%$
vs.\ $7.69\%$, and $7.26\%$ vs.\ $6.78\%$ for ConvNeXt-S, MIEW-ID, and
MegaDescriptor-L). At 50 distractors for the private jaguar data,
ConvNeXt-S and MegaDescriptor-L regress versus SphereFace2,
although both improve slightly over SphereFace2 in the full gallery.

\begin{figure*}[!t]
\centering
\includegraphics[width=0.76\textwidth]{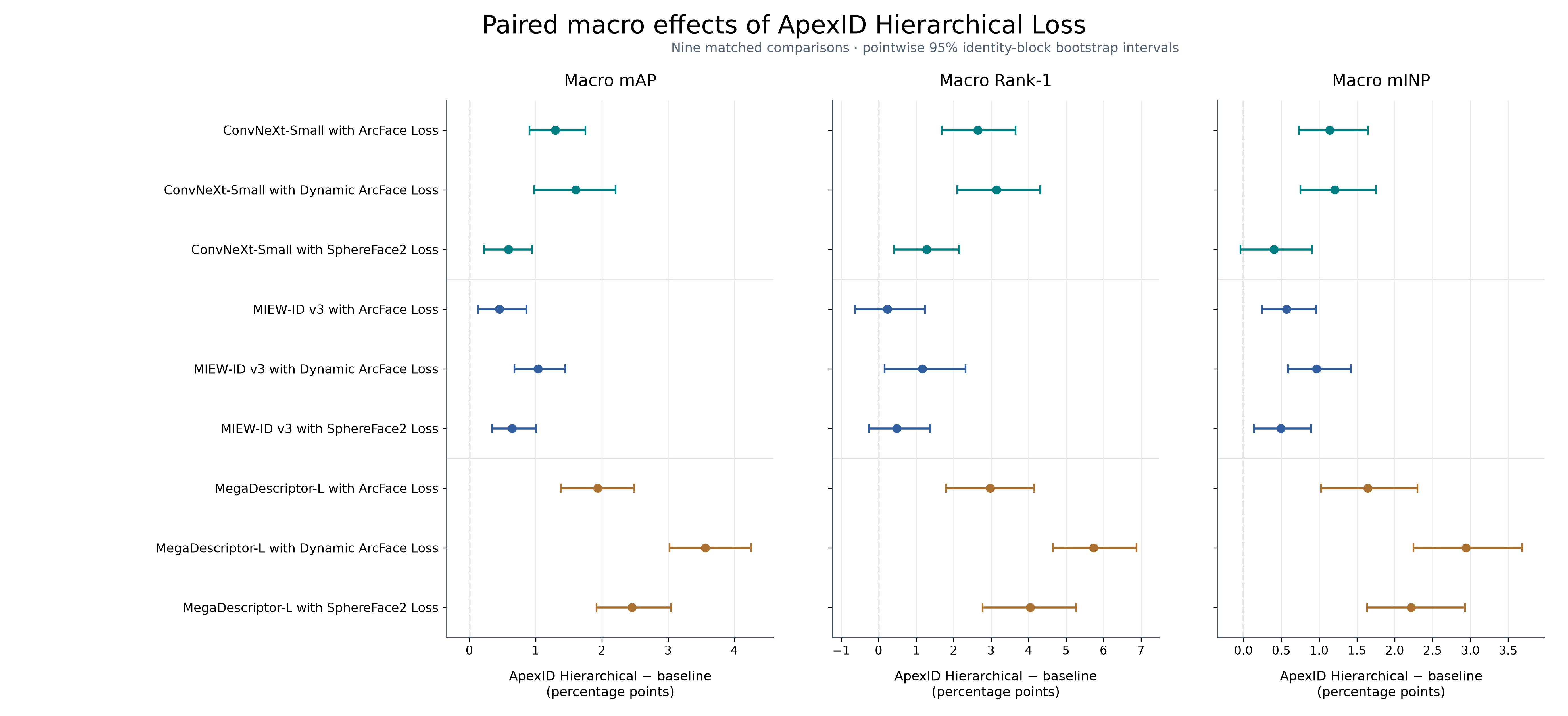}
\caption{Paired macro differences, ApexID Hierarchical minus baseline on
the same backbone, with pointwise 95\% identity-block bootstrap intervals.
Rank-1 and mINP intervals are secondary diagnostics.}
\label{fig:suppl:paired}

\vspace{0.75em}
\includegraphics[width=0.76\textwidth]{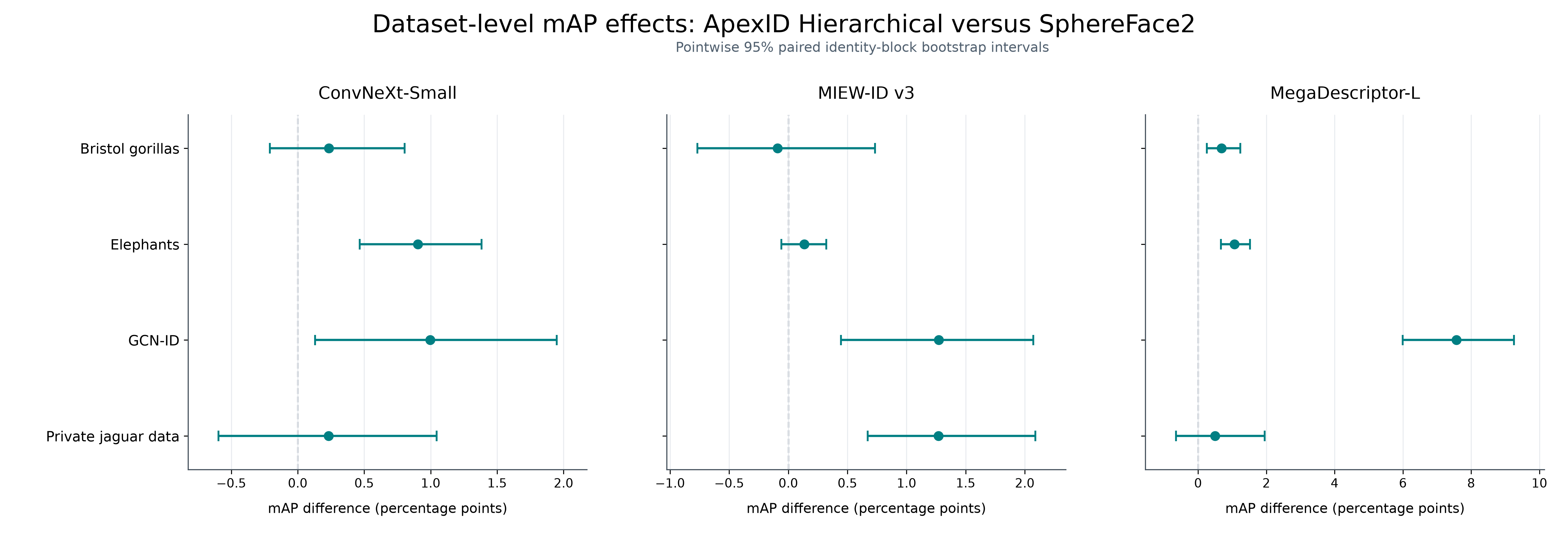}
\caption{Dataset-level mAP differences of ApexID Hierarchical versus SphereFace2.
Intervals are exploratory.}
\label{fig:suppl:dataset_effects}
\end{figure*}

\paragraph{Paired effects by dataset.}
Fig.~\ref{fig:suppl:paired} visualizes the nine macro contrasts.
Table~\ref{tab:suppl:dataset_effects} and Fig.~\ref{fig:suppl:dataset_effects}
break them down by dataset.
Against SphereFace2, the largest mAP gains occur on GCN-ID for all three
backbones: ConvNeXt-S ($+1.00$), MIEW-ID ($+1.27$, equal to its gain on the
private jaguar data), and MegaDescriptor-L ($+7.57$). mAP on the private
jaguar data increases for all three backbones ($+0.23$, $+1.27$, and
$+0.50$). Differences
are in percentage points. These subgroup intervals are 
exploratory.

\begin{figure*}[!t]
  \centering
  \includegraphics[width=0.46\textwidth]{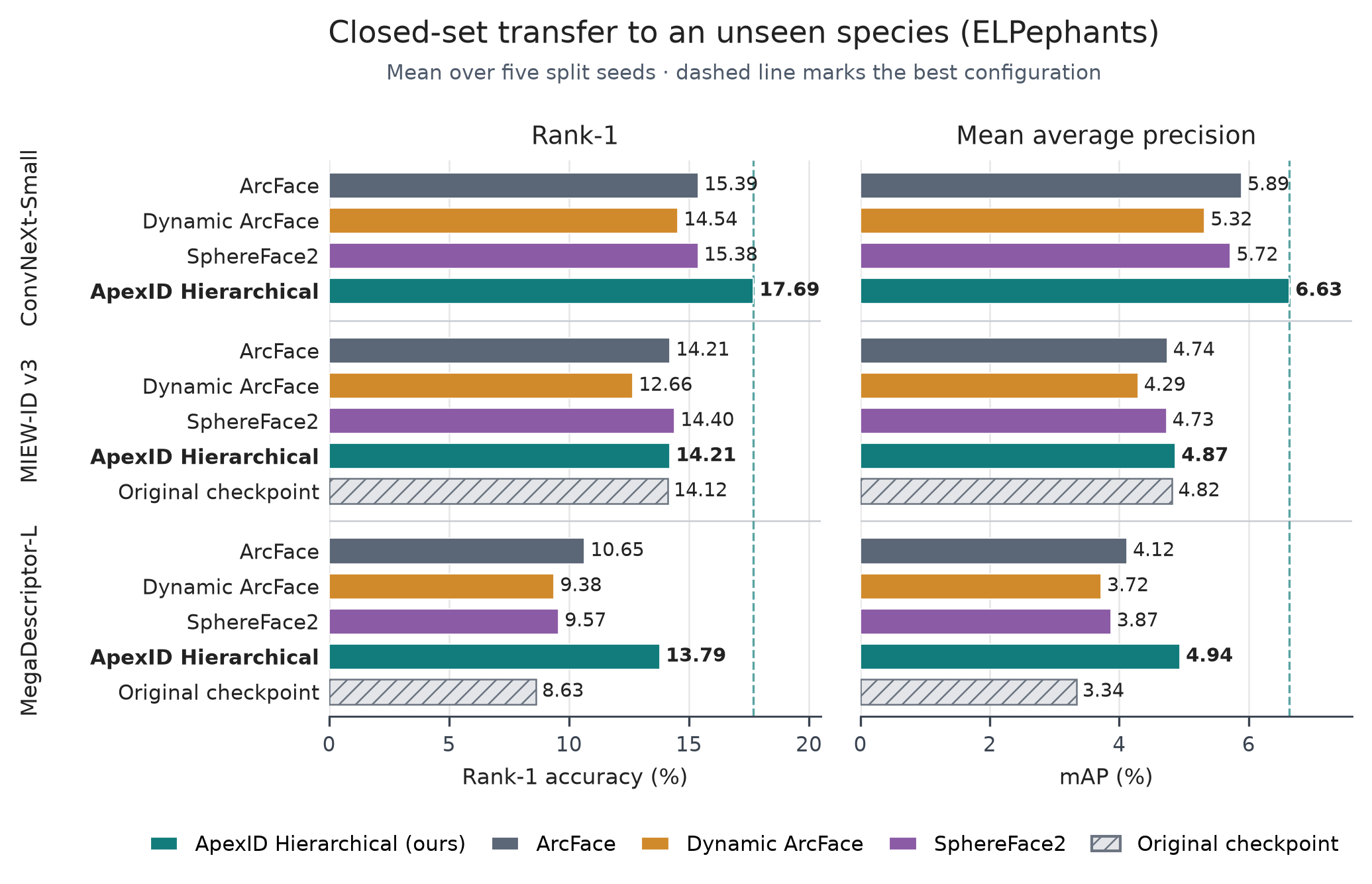}
  \caption{\textbf{Closed-set transfer to unseen species (ELPephants).}
  Rank-1 and mAP for fourteen configurations across three backbones.
  ApexID Hierarchical Loss (teal) gives the best result on ConvNeXt-S
  and MegaDescriptor-L. On MIEW-ID it has the highest mAP, slightly above the
  original checkpoint, while SphereFace2 has the highest Rank-1.  The dashed line marks the best
  configuration in each panel.}

  \label{fig:suppl:elephant}
\end{figure*}

\paragraph{Elephant transfer.}
The main transfer findings are reported in
Sec.~\ref{sec:results:retrieval} and Fig.~\ref{fig:suppl:elephant}. Table~\ref{tab:suppl:dataset_effects} and
Fig.~\ref{fig:suppl:dataset_effects} place these effects alongside the other
datasets.

\begin{table*}[!t]
\centering
\caption{Dataset-level paired differences in percentage points (ApexID Hierarchical Loss minus each
baseline on the same backbone, computed before rounding), averaged over five split seeds, with pointwise
95\% identity-block bootstrap intervals. }
\label{tab:suppl:dataset_effects}
\scriptsize
\setlength{\tabcolsep}{4pt}
\begin{tabular}{@{}l l l c c c@{}}
\toprule
\textbf{Backbone} & \textbf{vs.} & \textbf{Dataset} & \textbf{$\Delta$mAP} &
\textbf{$\Delta$R-1} & \textbf{$\Delta$mINP} \\
\midrule
ConvNeXt-S & ArcFace & Bristol & $-0.14$ {\tiny[$-0.62$, $+0.27$]} & $+0.64$ {\tiny[$-0.27$, $+1.78$]} & $-0.04$ {\tiny[$-0.09$, $+0.00$]} \\
 &  & Elephants & $\mathbf{+0.73}$ {\tiny[$+0.40$, $+1.10$]} & $\mathbf{+2.30}$ {\tiny[$+0.92$, $+3.67$]} & $+0.19$ {\tiny[$-0.06$, $+0.45$]} \\
 &  & GCN-ID & $\mathbf{+3.46}$ {\tiny[$+2.51$, $+4.48$]} & $\mathbf{+2.45}$ {\tiny[$+0.53$, $+4.26$]} & $\mathbf{+3.72}$ {\tiny[$+2.20$, $+5.43$]} \\
 &  & Private jaguar data & $\mathbf{+1.13}$ {\tiny[$+0.24$, $+2.47$]} & $\mathbf{+5.20}$ {\tiny[$+2.31$, $+8.28$]} & $\mathbf{+0.70}$ {\tiny[$+0.02$, $+1.87$]} \\
 & Dynamic ArcFace & Bristol & $+0.46$ {\tiny[$-0.20$, $+1.21$]} & $\mathbf{+1.24}$ {\tiny[$+0.23$, $+2.10$]} & $-0.03$ {\tiny[$-0.07$, $+0.01$]} \\
 &  & Elephants & $\mathbf{+1.31}$ {\tiny[$+0.87$, $+1.80$]} & $\mathbf{+3.14}$ {\tiny[$+1.71$, $+4.40$]} & $\mathbf{+0.42}$ {\tiny[$+0.09$, $+0.83$]} \\
 &  & GCN-ID & $\mathbf{+3.61}$ {\tiny[$+2.51$, $+4.62$]} & $\mathbf{+4.10}$ {\tiny[$+2.14$, $+6.09$]} & $\mathbf{+3.63}$ {\tiny[$+2.03$, $+5.47$]} \\
 &  & Private jaguar data & $+1.05$ {\tiny[$-0.87$, $+2.96$]} & $\mathbf{+4.09}$ {\tiny[$+0.83$, $+8.00$]} & $\mathbf{+0.81}$ {\tiny[$+0.18$, $+1.84$]} \\
 & SphereFace2 & Bristol & $+0.24$ {\tiny[$-0.21$, $+0.80$]} & $\mathbf{+0.73}$ {\tiny[$+0.20$, $+1.36$]} & $-0.01$ {\tiny[$-0.02$, $+0.00$]} \\
 &  & Elephants & $\mathbf{+0.90}$ {\tiny[$+0.47$, $+1.38$]} & $\mathbf{+2.30}$ {\tiny[$+0.94$, $+3.72$]} & $+0.31$ {\tiny[$-0.01$, $+0.71$]} \\
 &  & GCN-ID & $\mathbf{+1.00}$ {\tiny[$+0.13$, $+1.95$]} & $+0.26$ {\tiny[$-1.59$, $+2.02$]} & $+0.79$ {\tiny[$-0.77$, $+2.33$]} \\
 &  & Private jaguar data & $+0.23$ {\tiny[$-0.60$, $+1.04$]} & $+1.85$ {\tiny[$-1.16$, $+4.28$]} & $+0.52$ {\tiny[$-0.13$, $+1.74$]} \\
\midrule
MIEW-ID v3 & ArcFace & Bristol & $-0.47$ {\tiny[$-1.38$, $+0.67$]} & $+1.06$ {\tiny[$-0.37$, $+2.86$]} & $-0.01$ {\tiny[$-0.02$, $+0.00$]} \\
 &  & Elephants & $+0.13$ {\tiny[$-0.20$, $+0.50$]} & $-0.00$ {\tiny[$-1.43$, $+1.40$]} & $+0.12$ {\tiny[$-0.11$, $+0.47$]} \\
 &  & GCN-ID & $\mathbf{+1.14}$ {\tiny[$+0.38$, $+1.94$]} & $+0.27$ {\tiny[$-1.44$, $+2.23$]} & $\mathbf{+1.62}$ {\tiny[$+0.45$, $+3.08$]} \\
 &  & Private jaguar data & $\mathbf{+1.01}$ {\tiny[$+0.36$, $+1.95$]} & $-0.39$ {\tiny[$-2.43$, $+2.30$]} & $\mathbf{+0.54}$ {\tiny[$+0.12$, $+1.23$]} \\
 & Dynamic ArcFace & Bristol & $+0.39$ {\tiny[$-0.21$, $+1.32$]} & $+1.20$ {\tiny[$-0.18$, $+2.88$]} & $-0.00$ {\tiny[$-0.01$, $+0.01$]} \\
 &  & Elephants & $\mathbf{+0.58}$ {\tiny[$+0.26$, $+1.00$]} & $\mathbf{+1.55}$ {\tiny[$+0.32$, $+2.81$]} & $+0.16$ {\tiny[$-0.07$, $+0.52$]} \\
 &  & GCN-ID & $\mathbf{+2.01}$ {\tiny[$+1.09$, $+2.99$]} & $+1.60$ {\tiny[$-0.79$, $+4.97$]} & $\mathbf{+3.18}$ {\tiny[$+1.72$, $+4.80$]} \\
 &  & Private jaguar data & $\mathbf{+1.16}$ {\tiny[$+0.46$, $+2.21$]} & $+0.32$ {\tiny[$-1.85$, $+3.27$]} & $\mathbf{+0.53}$ {\tiny[$+0.08$, $+1.24$]} \\
 & SphereFace2 & Bristol & $-0.09$ {\tiny[$-0.77$, $+0.73$]} & $-0.04$ {\tiny[$-0.62$, $+0.65$]} & $-0.00$ {\tiny[$-0.01$, $+0.01$]} \\
 &  & Elephants & $+0.14$ {\tiny[$-0.06$, $+0.32$]} & $-0.19$ {\tiny[$-1.34$, $+0.90$]} & $+0.04$ {\tiny[$-0.02$, $+0.11$]} \\
 &  & GCN-ID & $\mathbf{+1.27}$ {\tiny[$+0.45$, $+2.07$]} & $+0.18$ {\tiny[$-1.47$, $+1.59$]} & $\mathbf{+1.40}$ {\tiny[$+0.13$, $+2.79$]} \\
 &  & Private jaguar data & $\mathbf{+1.27}$ {\tiny[$+0.67$, $+2.09$]} & $+1.99$ {\tiny[$-0.09$, $+4.77$]} & $\mathbf{+0.54}$ {\tiny[$+0.13$, $+1.17$]} \\
\midrule
MegaDescriptor-L & ArcFace & Bristol & $\mathbf{+0.78}$ {\tiny[$+0.11$, $+1.36$]} & $\mathbf{+2.95}$ {\tiny[$+1.40$, $+4.34$]} & $+0.01$ {\tiny[$-0.01$, $+0.03$]} \\
 &  & Elephants & $\mathbf{+0.82}$ {\tiny[$+0.33$, $+1.27$]} & $\mathbf{+3.15}$ {\tiny[$+1.72$, $+4.64$]} & $+0.04$ {\tiny[$-0.37$, $+0.39$]} \\
 &  & GCN-ID & $\mathbf{+5.59}$ {\tiny[$+3.93$, $+7.32$]} & $\mathbf{+4.53}$ {\tiny[$+1.81$, $+7.29$]} & $\mathbf{+6.31}$ {\tiny[$+4.08$, $+8.72$]} \\
 &  & Private jaguar data & $+0.55$ {\tiny[$-0.47$, $+1.81$]} & $+1.28$ {\tiny[$-1.86$, $+4.42$]} & $+0.21$ {\tiny[$-0.60$, $+1.11$]} \\
 & Dynamic ArcFace & Bristol & $\mathbf{+1.21}$ {\tiny[$+0.50$, $+1.93$]} & $\mathbf{+4.85}$ {\tiny[$+3.82$, $+6.03$]} & $+0.02$ {\tiny[$-0.01$, $+0.04$]} \\
 &  & Elephants & $\mathbf{+1.21}$ {\tiny[$+0.85$, $+1.63$]} & $\mathbf{+4.41}$ {\tiny[$+3.01$, $+5.94$]} & $\mathbf{+0.21}$ {\tiny[$+0.02$, $+0.47$]} \\
 &  & GCN-ID & $\mathbf{+10.25}$ {\tiny[$+8.64$, $+12.14$]} & $\mathbf{+8.82}$ {\tiny[$+6.24$, $+11.50$]} & $\mathbf{+10.53}$ {\tiny[$+7.93$, $+13.32$]} \\
 &  & Private jaguar data & $\mathbf{+1.57}$ {\tiny[$+0.36$, $+3.45$]} & $\mathbf{+4.86}$ {\tiny[$+1.95$, $+7.98$]} & $\mathbf{+1.02}$ {\tiny[$+0.23$, $+2.27$]} \\
 & SphereFace2 & Bristol & $\mathbf{+0.69}$ {\tiny[$+0.26$, $+1.24$]} & $\mathbf{+3.87}$ {\tiny[$+2.73$, $+5.15$]} & $+0.02$ {\tiny[$-0.02$, $+0.04$]} \\
 &  & Elephants & $\mathbf{+1.07}$ {\tiny[$+0.67$, $+1.52$]} & $\mathbf{+4.22}$ {\tiny[$+2.65$, $+5.97$]} & $+0.05$ {\tiny[$-0.27$, $+0.38$]} \\
 &  & GCN-ID & $\mathbf{+7.57}$ {\tiny[$+5.99$, $+9.26$]} & $\mathbf{+5.74}$ {\tiny[$+2.99$, $+8.79$]} & $\mathbf{+8.02}$ {\tiny[$+5.87$, $+10.55$]} \\
 &  & Private jaguar data & $+0.50$ {\tiny[$-0.65$, $+1.95$]} & $+2.35$ {\tiny[$-1.26$, $+5.72$]} & $+0.78$ {\tiny[$-0.01$, $+1.94$]} \\
\bottomrule
\end{tabular}
\end{table*}

\begin{table*}[!t]
\centering
\caption{Representation diagnostics on matched image samples comparing SphereFace2 with
ApexID Hierarchical Loss for each backbone. AUROC measures same-category verification (\%).
Boldface indicates that ApexID Hierarchical Loss exceeds SphereFace2 before rounding.}
\label{tab:suppl:representation}
\scriptsize
\setlength{\tabcolsep}{4pt}
\begin{tabular}{@{}l l rr rr rr rr@{}}
\toprule
& & \multicolumn{2}{c}{\textbf{AUROC}} &
\multicolumn{2}{c}{\textbf{Silhouette}} &
\multicolumn{2}{c}{\textbf{Fisher ratio}} &
\multicolumn{2}{c}{\textbf{Effective rank}} \\
\cmidrule(lr){3-4} \cmidrule(lr){5-6}
\cmidrule(lr){7-8} \cmidrule(lr){9-10}
\textbf{Backbone} & \textbf{Dataset} &
SphereFace2 & Ours & SphereFace2 & Ours &
SphereFace2 & Ours & SphereFace2 & Ours \\
\midrule
ConvNeXt-S & Bristol & $60.09$ & $59.92$ & $-0.0263$ & $-0.0314$ & $0.192$ & $\mathbf{0.192}$ & $36.3$ & $\mathbf{36.4}$ \\
 & Elephants & $62.29$ & $\mathbf{63.69}$ & $-0.1802$ & $\mathbf{-0.1574}$ & $0.427$ & $\mathbf{0.433}$ & $137.4$ & $\mathbf{139.0}$ \\
 & GCN-ID & $91.88$ & $91.68$ & $0.1904$ & $\mathbf{0.1971}$ & $1.199$ & $\mathbf{1.220}$ & $132.7$ & $131.6$ \\
 & \shortstack{Private jaguar data} & $75.49$ & $\mathbf{75.92}$ & $-0.0293$ & $\mathbf{-0.0290}$ & $0.616$ & $\mathbf{0.635}$ & $97.2$ & $92.4$ \\
\midrule
MIEW-ID v3 & Bristol & $60.73$ & $60.42$ & $-0.0109$ & $-0.0187$ & $0.166$ & $\mathbf{0.174}$ & $47.7$ & $44.9$ \\
 & Elephants & $60.08$ & $\mathbf{60.46}$ & $-0.1278$ & $\mathbf{-0.1209}$ & $0.341$ & $\mathbf{0.344}$ & $380.8$ & $374.6$ \\
 & GCN-ID & $92.68$ & $\mathbf{92.86}$ & $0.2348$ & $\mathbf{0.2478}$ & $1.143$ & $\mathbf{1.206}$ & $283.9$ & $264.4$ \\
 & \shortstack{Private jaguar data} & $72.41$ & $\mathbf{72.84}$ & $-0.0425$ & $-0.0598$ & $0.504$ & $\mathbf{0.555}$ & $137.2$ & $119.4$ \\
\midrule
MegaDescriptor-L & Bristol & $58.41$ & $\mathbf{58.81}$ & $-0.0199$ & $-0.0242$ & $0.163$ & $\mathbf{0.166}$ & $44.4$ & $42.9$ \\
 & Elephants & $57.34$ & $\mathbf{59.69}$ & $-0.2806$ & $\mathbf{-0.2146}$ & $0.366$ & $\mathbf{0.385}$ & $138.1$ & $\mathbf{154.2}$ \\
 & GCN-ID & $91.26$ & $\mathbf{93.02}$ & $0.1298$ & $\mathbf{0.1685}$ & $1.025$ & $\mathbf{1.120}$ & $194.5$ & $182.3$ \\
 & \shortstack{Private jaguar data} & $73.65$ & $\mathbf{75.93}$ & $-0.0664$ & $\mathbf{-0.0529}$ & $0.528$ & $\mathbf{0.603}$ & $128.5$ & $108.4$ \\
\bottomrule
\end{tabular}
\end{table*}

\paragraph{Representation and verification.}
Table~\ref{tab:suppl:representation} expands
Sec.~\ref{sec:results:representation} using matched samples of 56 Bristol,
1,500 elephant, 1,125 GCN-ID, and 289 private jaguar images.
Table~\ref{tab:suppl:verification} gives low-FAR verification. As a further
example of disagreement between geometry and retrieval, replacing SphereFace2
with ApexID Hierarchical decreases MegaDescriptor-L's GCN-ID effective rank from
$194.5$ to $182.3$ while increasing mAP by $7.57$ percentage points.

\begin{table*}[!t]
\centering
\caption{Same-category verification metrics (percentages): AUROC, equal error rate
(EER; lower is better), and true-accept rate (TAR) at 1\% and 0.1\% false
accept rates (FAR). Thresholds are derived from evaluation scores; these are
descriptive operating points. Boldface marks the best value within each
backbone family, including ties at the reported precision.}
\label{tab:suppl:verification}
\tiny
\setlength{\tabcolsep}{1.6pt}
\begin{tabular}{@{}l l rrrr rrrr rrrr rrrr@{}}
\toprule
& & \multicolumn{4}{c}{\textbf{Bristol}} &
\multicolumn{4}{c}{\textbf{Elephants}} &
\multicolumn{4}{c}{\textbf{GCN-ID}} &
\multicolumn{4}{c}{\textbf{\shortstack{Private\\jaguar data}}} \\
\cmidrule(lr){3-6} \cmidrule(lr){7-10}
\cmidrule(lr){11-14} \cmidrule(lr){15-18}
\textbf{Backbone} & \textbf{Loss} &
AUC & EER & T@1 & T@.1 &
AUC & EER & T@1 & T@.1 &
AUC & EER & T@1 & T@.1 &
AUC & EER & T@1 & T@.1 \\
\midrule
ConvNeXt-S & ArcFace & \textbf{60.81} & \textbf{41.80} & \textbf{5.40} & 1.12 & 63.10 & 40.90 & 5.69 & 1.66 & 91.05 & 15.66 & 61.11 & 44.19 & 75.15 & \textbf{31.56} & 19.59 & 5.67 \\
ConvNeXt-S & Dynamic ArcFace & 60.19 & 42.56 & 4.96 & 1.21 & 62.08 & 41.18 & 5.09 & 1.35 & 91.12 & 15.79 & 60.82 & 44.69 & 75.17 & 32.22 & 17.88 & 5.77 \\
ConvNeXt-S & SphereFace2 & 60.09 & 42.88 & 4.82 & \textbf{1.29} & 62.29 & 41.84 & 6.02 & 1.81 & \textbf{91.88} & \textbf{14.34} & 62.23 & 46.50 & 75.49 & 32.71 & \textbf{21.64} & 4.73 \\
ConvNeXt-S & ApexID Hierarchical Loss & 59.92 & 43.25 & 4.69 & 1.25 & \textbf{63.69} & \textbf{40.73} & \textbf{6.89} & \textbf{2.32} & 91.68 & 14.78 & \textbf{62.52} & \textbf{47.90} & \textbf{75.92} & 32.20 & 21.00 & \textbf{6.30} \\
\midrule
MIEW-ID v3 & ArcFace & \textbf{61.70} & \textbf{41.60} & \textbf{7.54} & \textbf{3.35} & 60.41 & \textbf{42.99} & 4.76 & 1.46 & \textbf{93.04} & 13.04 & \textbf{74.78} & 61.35 & 72.55 & 33.90 & 18.47 & 5.88 \\
MIEW-ID v3 & Dynamic ArcFace & 60.69 & 42.41 & 6.52 & 2.77 & 59.40 & 43.71 & 4.24 & 1.31 & 92.67 & 13.11 & 72.90 & 59.00 & \textbf{73.00} & 33.45 & 17.89 & 7.05 \\
MIEW-ID v3 & SphereFace2 & 60.73 & 42.97 & 6.70 & 2.90 & 60.08 & 43.08 & 5.04 & 1.50 & 92.68 & 13.35 & 73.68 & 59.80 & 72.41 & 34.31 & 17.20 & 6.18 \\
MIEW-ID v3 & ApexID Hierarchical Loss & 60.42 & 42.65 & 6.61 & 2.77 & \textbf{60.46} & 43.09 & \textbf{5.56} & \textbf{1.60} & 92.86 & \textbf{13.03} & 74.60 & \textbf{63.02} & 72.84 & \textbf{33.02} & \textbf{19.39} & \textbf{8.09} \\
\midrule
MegaDescriptor-L & ArcFace & 58.44 & 44.67 & 4.87 & 0.58 & 58.54 & 43.68 & 4.30 & 1.23 & 91.98 & 14.74 & 58.65 & 34.43 & 74.42 & 31.62 & 12.90 & 2.28 \\
MegaDescriptor-L & Dynamic ArcFace & 58.18 & 44.83 & 4.82 & 0.94 & 57.66 & 44.56 & 3.49 & 1.08 & 90.92 & 15.96 & 52.78 & 29.89 & 73.30 & 32.48 & 16.78 & 1.96 \\
MegaDescriptor-L & SphereFace2 & 58.41 & 45.04 & 5.04 & 0.94 & 57.34 & 44.88 & 2.89 & 0.94 & 91.26 & 15.58 & 53.35 & 32.18 & 73.65 & 32.02 & 16.51 & 2.22 \\
MegaDescriptor-L & ApexID Hierarchical Loss & \textbf{58.81} & \textbf{44.00} & \textbf{5.49} & \textbf{2.14} & \textbf{59.69} & \textbf{43.48} & \textbf{4.82} & \textbf{1.37} & \textbf{93.02} & \textbf{13.26} & \textbf{64.49} & \textbf{42.38} & \textbf{75.93} & \textbf{30.72} & \textbf{19.61} & \textbf{3.55} \\
\bottomrule
\end{tabular}
\end{table*}

\paragraph{Identity-level effects.}
Table~\ref{tab:suppl:identity_winloss} counts identities whose AP improves
or worsens relative to SphereFace2.
Average gains do not hold for every animal: on GCN-ID, MIEW-ID improves
95 identities but worsens 60, and on the private jaguar data ConvNeXt-S
improves 31 and worsens 19, and MegaDescriptor-L improves 31 and worsens 20.
On Bristol, ConvNeXt-S improves only 3 of the 7 identities.
Reliable camera, session, and side metadata were unavailable.
\newpage
\begin{table*}[!t]
\centering
\caption{Number of identities for which average precision (AP) improves, ties, or
worsens under ApexID Hierarchical Loss relative to SphereFace2 on the same backbone.}
\label{tab:suppl:identity_winloss}
\scriptsize
\setlength{\tabcolsep}{4pt}
\begin{tabular}{@{}l rrr rrr rrr rrr@{}}
\toprule
& \multicolumn{3}{c}{\textbf{Bristol}} &
\multicolumn{3}{c}{\textbf{Elephants}} &
\multicolumn{3}{c}{\textbf{GCN-ID}} &
\multicolumn{3}{c}{\textbf{Private jaguar data}} \\
\cmidrule(lr){2-4} \cmidrule(lr){5-7}
\cmidrule(lr){8-10} \cmidrule(lr){11-13}
\textbf{Backbone} &
Better & Tie & Worse & Better & Tie & Worse &
Better & Tie & Worse & Better & Tie & Worse \\
\midrule
ConvNeXt-S & 3 & 0 & 4 & 182 & 0 & 90 & 105 & 30 & 72 & 31 & 1 & 19 \\
MIEW-ID v3 & 4 & 0 & 3 & 156 & 0 & 116 & 95 & 52 & 60 & 35 & 1 & 15 \\
MegaDescriptor-L & 6 & 0 & 1 & 179 & 0 & 93 & 157 & 7 & 43 & 31 & 0 & 20 \\
\bottomrule
\end{tabular}
\end{table*}

\FloatBarrier

\newcommand{\scorecamimage}[1]{\includegraphics[width=0.235\textwidth]{figures/scorecam_ordered/#1}}
\newcommand{\scorecamrow}[4]{#1 & \scorecamimage{#2} & \scorecamimage{#3} & \scorecamimage{#4} \\}
\newcommand{\scorecamgrid}[1]{%
\centering
\scriptsize
\setlength{\tabcolsep}{2pt}
\begin{tabular}{@{}lccc@{}}
\toprule
\textbf{Objective} & \textbf{ConvNeXt-S} & \textbf{MIEW-ID v3} &
\textbf{MegaDescriptor-L} \\
\midrule
\scorecamrow{ArcFace}{convnext_af_#1.png}{miewid_af_#1.png}{megadescriptor_af_#1.png}
\scorecamrow{Dynamic ArcFace}{convnext_daf_#1.png}{miewid_daf_#1.png}{megadescriptor_daf_#1.png}
\scorecamrow{SphereFace2}{convnext_sf2_#1.png}{miewid_sf2_#1.png}{megadescriptor_sf2_#1.png}
\scorecamrow{ApexID Hierarchical}{convnext_scsf2_#1.png}{miewid_scsf2_#1.png}{megadescriptor_scsf2_#1.png}
\bottomrule
\end{tabular}}

\twocolumn[{\vbox to 0.97\textheight{%
\section{ScoreCAM Visualizations}
\label{sec:suppl:scorecam}

Each grid presents one ScoreCAM example for every backbone--objective pair
for a given dataset. Columns correspond to backbones, and rows to training
objectives. Within each group of three images, the panels show the overlay, heatmap, and
original image from left to right.

\vfill
\begin{minipage}{\textwidth}
\scorecamgrid{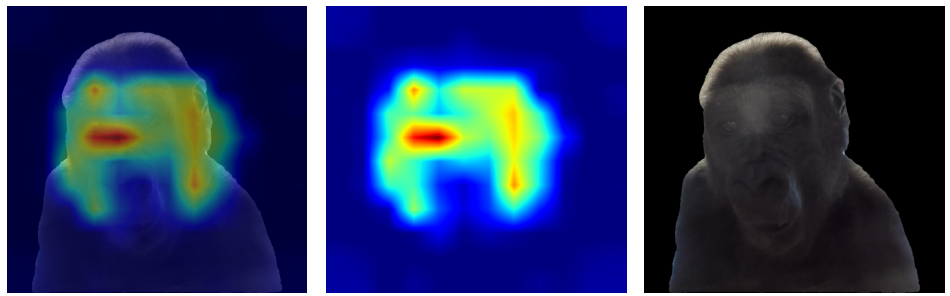}
\captionof{figure}{ScoreCAM examples from the Bristol gorillas dataset, arranged by
objective (rows) and backbone (columns).}
\label{fig:suppl:scorecam:bristol}
\end{minipage}

\vfill
\begin{minipage}{\textwidth}
\scorecamgrid{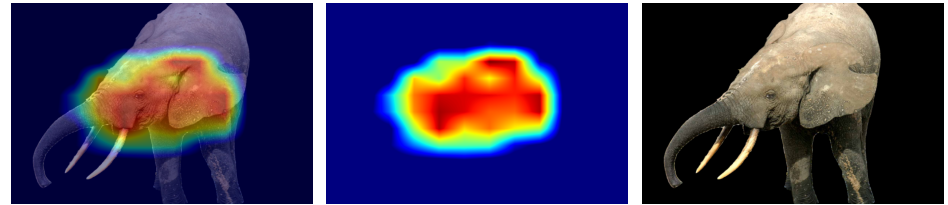}
\captionof{figure}{ScoreCAM examples from ELPephants, arranged by objective (rows) and
backbone (columns).}
\label{fig:suppl:scorecam:elephant}
\end{minipage}
\vfill
}}]

\twocolumn[{%
\begin{minipage}{\textwidth}
\scorecamgrid{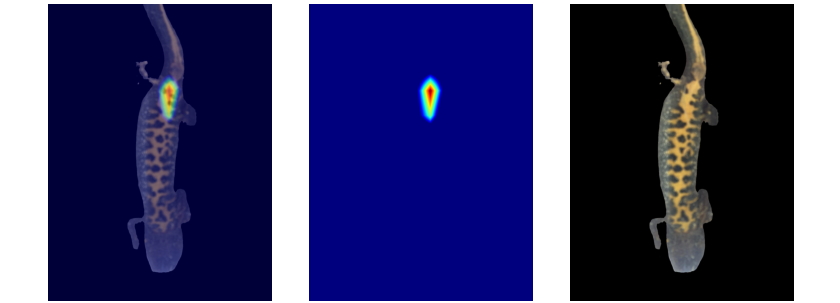}
\captionof{figure}{ScoreCAM examples from GCN-ID, arranged by objective (rows) and
backbone (columns).}
\label{fig:suppl:scorecam:gcn}
\end{minipage}

\vspace{1.2em}
\begin{minipage}{\textwidth}
\scorecamgrid{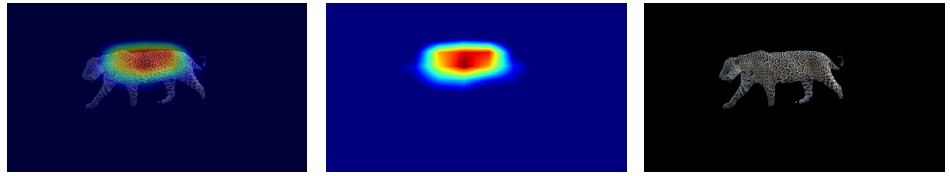}
\captionof{figure}{ScoreCAM examples from the private jaguar dataset, arranged by
objective (rows) and backbone (columns).}
\label{fig:suppl:scorecam:jaguar}
\end{minipage}
\vspace{1.5em}
}]

\section{Per-dataset Open-Set Analysis}
\label{sec:suppl:per_dataset_open_set}

Table~\ref{tab:open_world_results} breaks down open-set discovery and
known-identity attachment by dataset for all fourteen embeddings.
We calibrate similarities using a fixed, dataset-stratified cohort of 512
labelled images (seed~$42$), drawn from canonical catalogues that include the
external datasets. We use the same images for every model and fit the
mappings separately for each descriptor. For gallery datasets,
Optuna~\cite{akiba2019optuna} selects reranking, clustering, and attachment
parameters per model--dataset pair by maximizing
$0.6\,\mathrm{ARI}+0.4\,\mathrm{V}$, where $\mathrm{V}$ is
V-measure~\cite{rosenberg2007v}. Selected configurations are fixed before
final clustering. For gallery-free jaguar discovery, we use fixed defaults and
do not tune on query labels. Calibration samples are part of the gallery,
and query labels used for selection remain in the reported pool; these
results therefore characterize dataset-adapted association and discovery.
\clearpage
\begin{table*}[p]
\centering
\caption{Per-dataset open-set discovery, novelty, and known-identity
attachment metrics (split seed~42). Values lie in $[0,1]$; dashes indicate
metrics not defined for gallery-free jaguar discovery. Boldface indicates
the best value within each backbone family for each dataset; ties at the
reported precision are also bold.}
\label{tab:open_world_results}
\scriptsize
\setlength{\tabcolsep}{1.2pt}
\begin{tabular}{@{}ll r r r r r r r r r r r@{}}
\toprule
\textbf{Embedding} & \textbf{Dataset} & \textbf{ARI} & \textbf{AMI} &
\textbf{V} & \textbf{B$^3$ F1} & \textbf{Pair F1} & \textbf{Nov.\ ARI} &
\textbf{Nov.\ B$^3$} & \textbf{BAKS} & \textbf{NormAcc} &
\textbf{Known att.} & \textbf{BAUS} \\
\midrule
ConvNeXt-S ArcFace & Bristol & 0.174 & \textbf{0.462} & \textbf{0.483} & \textbf{0.348} & 0.251 & 0.000 & 0.192 & 0.000 & 0.000 & 0.000 & \textbf{1.000} \\
 & Elephants & 0.242 & \textbf{0.393} & 0.890 & \textbf{0.600} & 0.244 & 0.233 & 0.447 & \textbf{0.007} & \textbf{0.087} & \textbf{0.005} & 0.999 \\
 & GCN-ID & 0.938 & 0.969 & 0.993 & 0.978 & 0.939 & \textbf{0.943} & \textbf{0.971} & 0.287 & 0.536 & 0.333 & \textbf{1.000} \\
 & Jaguar & 0.342 & 0.652 & \textbf{0.780} & 0.458 & 0.356 & 0.342 & 0.458 & {---} & {---} & {---} & \textbf{1.000} \\
ConvNeXt-S Dynamic ArcFace & Bristol & 0.167 & 0.444 & 0.466 & 0.344 & 0.247 & 0.000 & 0.194 & 0.000 & 0.000 & 0.000 & \textbf{1.000} \\
 & Elephants & 0.250 & 0.382 & 0.886 & 0.591 & 0.252 & 0.247 & 0.449 & 0.005 & 0.071 & 0.003 & \textbf{1.000} \\
 & GCN-ID & 0.940 & 0.971 & \textbf{0.994} & 0.979 & 0.940 & \textbf{0.943} & \textbf{0.971} & 0.910 & 0.942 & 0.910 & 0.976 \\
 & Jaguar & 0.387 & 0.658 & 0.778 & 0.473 & 0.402 & 0.387 & 0.473 & {---} & {---} & {---} & \textbf{1.000} \\
ConvNeXt-S SphereFace2 & Bristol & 0.175 & 0.444 & 0.462 & 0.319 & 0.246 & 0.000 & 0.193 & 0.000 & 0.000 & 0.000 & \textbf{1.000} \\
 & Elephants & 0.249 & 0.376 & 0.872 & 0.578 & 0.252 & \textbf{0.269} & \textbf{0.465} & 0.005 & 0.071 & 0.003 & \textbf{1.000} \\
 & GCN-ID & 0.940 & 0.971 & \textbf{0.994} & 0.979 & 0.940 & \textbf{0.943} & \textbf{0.971} & 0.910 & 0.942 & 0.910 & 0.976 \\
 & Jaguar & \textbf{0.398} & \textbf{0.663} & \textbf{0.780} & \textbf{0.475} & \textbf{0.413} & \textbf{0.398} & \textbf{0.475} & {---} & {---} & {---} & \textbf{1.000} \\
ConvNeXt-S ApexID Hierarchical Loss & Bristol & \textbf{0.180} & 0.450 & 0.465 & 0.333 & \textbf{0.266} & 0.000 & \textbf{0.234} & 0.000 & 0.000 & 0.000 & \textbf{1.000} \\
 & Elephants & \textbf{0.267} & 0.380 & \textbf{0.893} & 0.597 & \textbf{0.269} & 0.266 & 0.438 & 0.000 & 0.000 & 0.000 & \textbf{1.000} \\
 & GCN-ID & \textbf{0.941} & \textbf{0.973} & \textbf{0.994} & \textbf{0.980} & \textbf{0.941} & \textbf{0.943} & \textbf{0.971} & \textbf{0.958} & \textbf{0.967} & \textbf{0.958} & 0.976 \\
 & Jaguar & 0.339 & 0.651 & 0.779 & 0.455 & 0.354 & 0.339 & 0.455 & {---} & {---} & {---} & \textbf{1.000} \\
\midrule
MIEW-ID v3 ArcFace & Bristol & 0.171 & 0.447 & 0.472 & 0.326 & 0.247 & 0.000 & 0.199 & 0.073 & 0.270 & 0.064 & \textbf{0.997} \\
 & Elephants & 0.261 & 0.384 & 0.876 & 0.582 & 0.263 & 0.285 & \textbf{0.470} & 0.000 & 0.000 & 0.000 & \textbf{1.000} \\
 & GCN-ID & \textbf{0.940} & \textbf{0.971} & \textbf{0.994} & \textbf{0.979} & \textbf{0.940} & \textbf{0.943} & \textbf{0.971} & 0.886 & 0.929 & 0.889 & \textbf{0.976} \\
 & Jaguar & 0.361 & 0.628 & 0.759 & 0.439 & 0.376 & 0.361 & 0.439 & {---} & {---} & {---} & \textbf{1.000} \\
MIEW-ID v3 Dynamic ArcFace & Bristol & 0.196 & 0.428 & 0.439 & 0.360 & 0.316 & 0.000 & 0.352 & 0.054 & 0.231 & 0.051 & \textbf{0.997} \\
 & Elephants & 0.235 & 0.374 & 0.882 & 0.582 & 0.237 & 0.246 & 0.451 & 0.005 & 0.071 & 0.003 & \textbf{1.000} \\
 & GCN-ID & \textbf{0.940} & \textbf{0.971} & \textbf{0.994} & \textbf{0.979} & \textbf{0.940} & \textbf{0.943} & \textbf{0.971} & \textbf{0.934} & \textbf{0.954} & \textbf{0.931} & \textbf{0.976} \\
 & Jaguar & \textbf{0.406} & \textbf{0.635} & \textbf{0.762} & \textbf{0.451} & \textbf{0.421} & \textbf{0.406} & \textbf{0.451} & {---} & {---} & {---} & \textbf{1.000} \\
MIEW-ID v3 SphereFace2 & Bristol & 0.151 & 0.429 & \textbf{0.476} & 0.363 & 0.233 & 0.000 & 0.144 & \textbf{0.382} & 0.558 & 0.332 & 0.816 \\
 & Elephants & 0.252 & 0.381 & \textbf{0.888} & \textbf{0.595} & 0.255 & 0.258 & 0.450 & \textbf{0.014} & \textbf{0.118} & \textbf{0.011} & 0.987 \\
 & GCN-ID & \textbf{0.940} & \textbf{0.971} & \textbf{0.994} & \textbf{0.979} & \textbf{0.940} & \textbf{0.943} & \textbf{0.971} & 0.898 & 0.936 & 0.899 & \textbf{0.976} \\
 & Jaguar & 0.365 & 0.626 & 0.758 & 0.440 & 0.381 & 0.365 & 0.440 & {---} & {---} & {---} & \textbf{1.000} \\
MIEW-ID v3 ApexID Hierarchical Loss & Bristol & \textbf{0.203} & \textbf{0.450} & 0.462 & \textbf{0.379} & \textbf{0.325} & 0.000 & \textbf{0.389} & 0.234 & 0.373 & 0.185 & 0.594 \\
 & Elephants & 0.260 & 0.383 & 0.883 & 0.589 & 0.263 & 0.277 & 0.458 & 0.000 & 0.000 & 0.000 & \textbf{1.000} \\
 & GCN-ID & 0.937 & 0.968 & 0.993 & 0.978 & 0.938 & 0.940 & 0.968 & 0.904 & 0.939 & 0.905 & \textbf{0.976} \\
 & Jaguar & 0.317 & 0.621 & 0.752 & 0.427 & 0.333 & 0.317 & 0.427 & {---} & {---} & {---} & \textbf{1.000} \\
MIEW-ID v3 Pretrained & Bristol & 0.182 & 0.434 & 0.465 & 0.361 & 0.255 & 0.000 & 0.154 & 0.349 & \textbf{0.576} & \textbf{0.410} & 0.951 \\
 & Elephants & \textbf{0.285} & \textbf{0.397} & 0.886 & 0.591 & \textbf{0.287} & \textbf{0.306} & 0.469 & 0.005 & 0.071 & 0.003 & \textbf{1.000} \\
 & GCN-ID & \textbf{0.940} & \textbf{0.971} & \textbf{0.994} & \textbf{0.979} & \textbf{0.940} & \textbf{0.943} & \textbf{0.971} & 0.861 & 0.917 & 0.873 & \textbf{0.976} \\
 & Jaguar & 0.397 & 0.619 & 0.753 & 0.437 & 0.412 & 0.397 & 0.437 & {---} & {---} & {---} & \textbf{1.000} \\
\midrule
MegaDescriptor-L ArcFace & Bristol & 0.153 & 0.450 & \textbf{0.499} & 0.371 & 0.227 & 0.000 & 0.122 & 0.371 & \textbf{0.574} & 0.329 & 0.888 \\
 & Elephants & 0.214 & 0.349 & 0.884 & 0.582 & 0.216 & 0.205 & 0.419 & 0.000 & 0.000 & 0.000 & \textbf{1.000} \\
 & GCN-ID & 0.943 & 0.974 & 0.994 & 0.981 & \textbf{0.944} & \textbf{0.946} & \textbf{0.974} & 0.886 & 0.929 & 0.889 & \textbf{0.976} \\
 & Jaguar & 0.347 & 0.643 & 0.771 & 0.447 & 0.363 & 0.347 & 0.447 & {---} & {---} & {---} & \textbf{1.000} \\
MegaDescriptor-L Dynamic ArcFace & Bristol & \textbf{0.198} & 0.432 & 0.477 & \textbf{0.406} & \textbf{0.300} & 0.000 & 0.270 & \textbf{0.462} & 0.455 & \textbf{0.390} & 0.449 \\
 & Elephants & 0.211 & 0.340 & 0.887 & 0.582 & 0.213 & 0.201 & 0.412 & \textbf{0.005} & \textbf{0.071} & \textbf{0.003} & \textbf{1.000} \\
 & GCN-ID & 0.943 & 0.974 & 0.994 & 0.981 & \textbf{0.944} & \textbf{0.946} & \textbf{0.974} & \textbf{0.958} & \textbf{0.967} & 0.958 & \textbf{0.976} \\
 & Jaguar & \textbf{0.420} & 0.652 & 0.776 & 0.468 & \textbf{0.437} & \textbf{0.420} & 0.468 & {---} & {---} & {---} & \textbf{1.000} \\
MegaDescriptor-L SphereFace2 & Bristol & 0.150 & \textbf{0.455} & 0.490 & 0.267 & 0.199 & 0.000 & 0.133 & 0.000 & 0.000 & 0.000 & \textbf{1.000} \\
 & Elephants & 0.208 & 0.333 & 0.881 & 0.574 & 0.210 & 0.203 & 0.418 & \textbf{0.005} & \textbf{0.071} & \textbf{0.003} & \textbf{1.000} \\
 & GCN-ID & \textbf{0.944} & \textbf{0.975} & \textbf{0.995} & \textbf{0.982} & \textbf{0.944} & \textbf{0.946} & \textbf{0.974} & \textbf{0.958} & \textbf{0.967} & \textbf{0.963} & \textbf{0.976} \\
 & Jaguar & 0.340 & 0.641 & 0.771 & 0.445 & 0.355 & 0.340 & 0.445 & {---} & {---} & {---} & \textbf{1.000} \\
MegaDescriptor-L ApexID Hierarchical Loss & Bristol & 0.188 & 0.447 & 0.459 & 0.363 & 0.285 & 0.000 & \textbf{0.295} & 0.000 & 0.000 & 0.000 & \textbf{1.000} \\
 & Elephants & 0.224 & \textbf{0.361} & \textbf{0.889} & \textbf{0.593} & 0.226 & 0.209 & 0.425 & 0.000 & 0.000 & 0.000 & \textbf{1.000} \\
 & GCN-ID & 0.940 & 0.971 & 0.994 & 0.979 & 0.940 & 0.943 & 0.971 & 0.922 & 0.948 & 0.921 & \textbf{0.976} \\
 & Jaguar & 0.384 & \textbf{0.653} & \textbf{0.777} & \textbf{0.469} & 0.399 & 0.384 & \textbf{0.469} & {---} & {---} & {---} & \textbf{1.000} \\
MegaDescriptor-L Pretrained & Bristol & 0.175 & 0.414 & 0.426 & 0.325 & 0.273 & 0.000 & 0.259 & 0.000 & 0.000 & 0.000 & \textbf{1.000} \\
 & Elephants & \textbf{0.246} & 0.359 & 0.879 & 0.580 & \textbf{0.249} & \textbf{0.248} & \textbf{0.432} & 0.000 & 0.000 & 0.000 & \textbf{1.000} \\
 & GCN-ID & 0.943 & 0.974 & 0.994 & 0.981 & \textbf{0.944} & \textbf{0.946} & \textbf{0.974} & 0.952 & 0.964 & 0.952 & \textbf{0.976} \\
 & Jaguar & 0.348 & 0.621 & 0.757 & 0.435 & 0.363 & 0.348 & 0.435 & {---} & {---} & {---} & \textbf{1.000} \\
\bottomrule
\end{tabular}
\end{table*}

\clearpage
\paragraph{GCN-ID.}
Discovery scores are consistently high across embeddings (ARI
$0.937$--$0.944$; B$^3$ F1 $0.978$--$0.982$). The highest BAKS ($0.958$) is
shared by ConvNeXt-S ApexID Hierarchical, MegaDescriptor-L Dynamic ArcFace,
and MegaDescriptor-L SphereFace2 at the reported precision.

\paragraph{Elephants.}
ApexID Hierarchical has the highest ARI among the fine-tuned objectives
for ConvNeXt-S ($0.267$) and MegaDescriptor-L ($0.224$); on MIEW-ID, its ARI
($0.260$) is just below ArcFace ($0.261$). The unmodified MIEW-ID checkpoint
is highest overall ($0.285$). Known-identity attachment is close to zero for every
embedding (BAKS $\leq 0.014$).

\paragraph{Bristol.}
MegaDescriptor-L Dynamic ArcFace has the highest BAKS ($0.462$) but rejects fewer than
half of the novel queries (BAUS $0.449$), whereas the pretrained MIEW-ID
checkpoint has the highest NormAcc ($0.576$).

\paragraph{Private jaguar data.}
For gallery-free discovery, MegaDescriptor-L Dynamic ArcFace has the highest ARI
($0.420$), and ApexID Hierarchical has the lowest ARI among fine-tuned ConvNeXt-S and
MIEW-ID objectives ($0.339$ and $0.317$). Attachment metrics are not defined
in this setting.

\end{document}